\documentclass[12pt, a4paper, table]{article}

\usepackage[utf8]{inputenc} 
\usepackage{amsmath, amsthm, amssymb, amsfonts} 
\usepackage{lipsum} 

\usepackage{graphicx} 
\usepackage{float} 
\usepackage{caption} 
\usepackage{subcaption} 

\usepackage{tabularx} 
\usepackage{tabu} 
\usepackage{booktabs} 

\usepackage[nottoc,numbib]{tocbibind} 
\usepackage[margin = 2.5cm]{geometry} 
\usepackage{microtype} 
\usepackage{titlesec} 
\usepackage{titletoc} 
\usepackage{appendix} 
\usepackage{fancyhdr} 
\usepackage[shortlabels]{enumitem} 

\usepackage{hyperref} 
\usepackage[noabbrev, capitalise]{cleveref} 

\usepackage{xcolor} 

\usepackage{tikz} 
\usetikzlibrary{positioning} 
\usetikzlibrary{calc} 

\makeatletter
\newcommand{\gettikzxy}[3]{%
  \tikz@scan@one@point\pgfutil@firstofone#1\relax
  \edef#2{\the\pgf@x}%
  \edef#3{\the\pgf@y}%
}
\makeatother

\theoremstyle{definition}
\newtheorem{definition}{Definition}[section]
\usepackage{dsfont}

\newcommand\reporttitle{Unsupervised Domain Adaptation with Deep Neural-Networks}

\newcommand\reportsubtitle{
MSIAM M2, 2022-2023
}

\newcommand\reportauthors{
\\\\\\\\\\\\\
\Large Artem Bitiutskii \\
}

\newcommand\grouptutor{
\large Supervisors: Massih-Reza Amini and Marianne Clausel
}

\definecolor{Tue-red}{RGB}{199, 25, 24}

\hypersetup{
    colorlinks=true,
    linkcolor=Tue-red,
    urlcolor=Tue-red,
    citecolor = Tue-red
    }
\titleformat{\section}{\sffamily\color{Tue-red}\Large\bfseries}{\thesection\enskip\color{gray}\textbar\enskip}{0cm}{} 

\titleformat{\subsection}{\sffamily\color{Tue-red}\large\bfseries}{\thesubsection\enskip\color{gray}\textbar\enskip}{0cm}{} 

\titleformat{\subsubsection}{\sffamily\color{Tue-red}\bfseries}{\thesubsubsection\enskip\color{gray}\textbar\enskip}{0cm}{} 

\DeclareCaptionFont{Tue-red}{\color{Tue-red}}
\begin{document}

\begin{titlepage}

\centering

\begin{tikzpicture}

/
\node[opacity=0.3,inner sep=0pt,remember picture,overlay] at (4.5,-0.5){\includegraphics[width= 0.8 \textwidth]{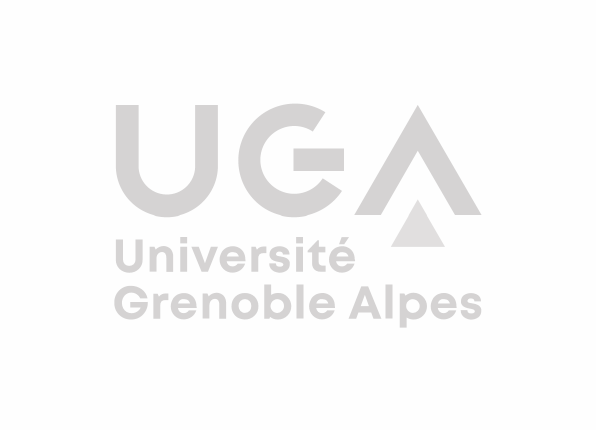}};

\node[inner sep=0pt] (logo) at (0,0)
    {\includegraphics[width=.25\textwidth]{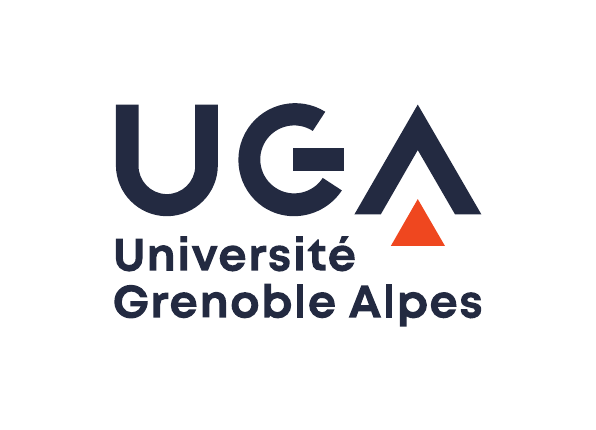}};
    
\node[text width = 0.5\textwidth, right = of logo](title){\sffamily\huge\reporttitle};

\node[text width = 0.5\textwidth, yshift = 0.75cm, below = of title](subtitle){\sffamily\Large \reportsubtitle};

\gettikzxy{(subtitle.south)}{\sffamily\subtitlex}{\subtitley}
\gettikzxy{(title.north)}{\titlex}{\titley}
\draw[line width=1mm, black]($(logo.east)!0.5!(title.west)$) +(0,\subtitley) -- +(0,\titley);

\end{tikzpicture}
\vspace{3cm}


\begin{table}[H]
\centering
\sffamily
\large
\begin{tabu} to 0.8\linewidth {cc}

\sffamily\reportauthors

\end{tabu}

\end{table}

\sffamily \grouptutor

\tikz[remember picture,overlay]\node[anchor=south,inner sep=0pt] at (current page.south) {\includegraphics[width=\paperwidth]{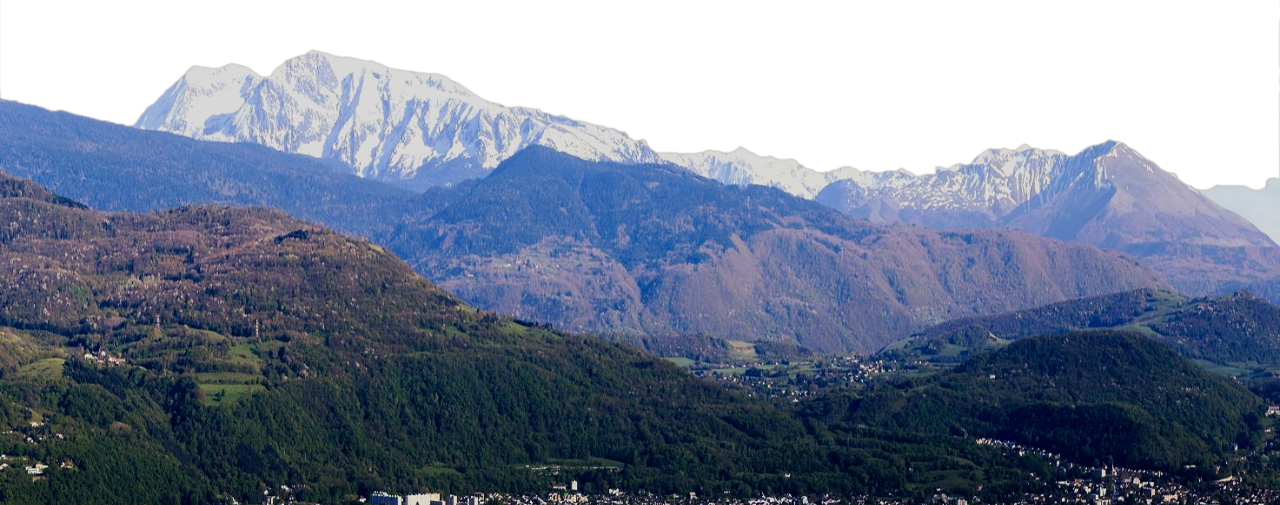}};

\mbox{}
\vfill
\sffamily \Large \textcolor{white}{Grenoble} \\

\end{titlepage}

\newpage

{\hypersetup{linkcolor=black} 
\tableofcontents\thispagestyle{empty}}
\newpage



\pagenumbering{arabic}
\section{Introduction} \label{section: introduction}

\emph{We start with necessary theory and motivation to understand some essential definitions and ideas associated with domain adaptation.}\\

\subsection{Theory}

Domain adaptation is a subfield of machine learning that deals with the problem of transferring knowledge learned from one domain to another related but different domain. In real-world scenarios, it is common to encounter situations where the data distribution of the target domain differs significantly from the source domain used to train a model. This can lead to a significant drop in the performance of the model on the target domain.\\

To understand clearly what domain adaptation is about, we should start with transfer learning. For this purpose, we should dig deeper into the theory. In these articles \cite{redko2020survey} and \cite{wilson2020survey}  you can find a high level overview of the theory that connects with domain adaptation. Let's start with the transfer learning definition and types that it consists of. 

\begin{definition}
(Transfer learning) We consider a source data distribution S called the source domain, and a target
data distribution $T$ called the target domain. Let $X_S$ × $Y_S$ be the source input and output spaces associated to $S$, and $X_T$ × $Y_T$ be the target input and output spaces associated to $T$. We use $S_X$ and $T_X$ to denote the marginal distributions of $X_S$ and $X_T$ , $t_S$ and $t_T$ to denote the source and target learning tasks depending on $Y_S$ and $Y_T$, respectively. Then, transfer learning aims to help to improve the learning of the target predictive function $f_T$ : $X_T \longrightarrow Y_T$ for $t_T$ using
knowledge gained from $S$ and $t_S$ , where $S = T$.
\end{definition}

According to these papers (\cite{redko2020survey}, \cite{wilson2020survey}), transfer learning algorithms can be classified into three categories based on the differences between the source and target tasks and domains: inductive, transductive, and unsupervised transfer learning. 
\begin{itemize}
    \item \textbf{Inductive transfer learning} involves using labeled data from the source domain to train a model for a different, but related, target task in the target domain. In this case, some labeled data from the target domain is required to fine-tune the model.

    \item \textbf{Transductive transfer learning}, on the other hand, refers to using both labeled data from the source domain and unlabeled data from the target domain to improve the model's performance on the target domain. In this case, the tasks remain the same while the domains are different.

    \item \textbf{Unsupervised transfer learning} involves adapting a model trained on the source task to perform well on a related, but different target task in the target domain, without any labeled data in either the source or target domains.
\end{itemize}

Domain adaptation is a type of transfer learning where the target task remains the same as the source task, but the domain differs (the second type -- transductive transfer learning). Depending on whether the feature spaces remain the same or differ, domain adaptation is categorized into homogeneous and heterogeneous domain adaptation. Machine learning techniques are commonly categorized based on the availability of labeled training data, such as supervised, semi-supervised, and unsupervised learning. However, domain adaptation assumes the availability of data from both the source and target domains, making it ambiguous to append one of these three terms to "domain adaptation". There are different ways how these terms can be applied to domain adaptation, but we use the same as in \cite{wilson2020survey}. 

\begin{itemize}
    \item \textbf{Unsupervised domain adaptation} refers to the case where both labeled source data and unlabeled target data are available
    \item \textbf{Semi-supervised domain adaptation} refers to the case where labeled source data and some labeled target data are available
    \item \textbf{Supervised domain adaptation} refers to the case where both labeled source and target data are available.
\end{itemize}

Unsupervised Domain adaptation could be applied to a wide range of tasks in NLP \cite{Amini00, Pessiot09}, in vision \cite{zhang2021survey} and in many other applications where assigning labels to examples is tedious or impossible.

This report is more focused on studying unsupervised domain adaptation with using deep neural-networks. Before move on to the practical part, it is important to discuss theoretical analysis and guarantees that can be used in fields associated with transfer learning. Thus, there are several methods that allow you to analyze the generalization gap in machine learning \cite{wang2018theoretical}. One of the most popular approaches is the model complexity approach, which estimates the generalization bound by measuring the complexity of the hypothesis set, such as Vapnik-Chervonenkis (VC) dimension and Rademacher complexity. Another approach is to use the stability of the supervised learning algorithm in relation to the datasets. Stability is a measure of how much a change in a data point in the training set can affect the output of the algorithm. Both of these approaches have been used to analyze the generalization bounds of transfer learning algorithms.\\

It is equally important to discuss distributions and what experts mean by shift when analyzing transfer learning algorithms. Distribution refers to the set of all possible values of a random variable, and a shift refers to a change in the distribution of the data between the source and target domains. Understanding the shift in the distribution of the data is crucial in developing effective transfer learning algorithms, as it enables the selection of appropriate techniques for adapting the model to the target domain. \\

Unsupervised domain adaptation (UDA) is a type of supervised learning that involves training a model using labeled source data and applying it to unlabeled target data, where the distributions of the two domains differ. Let the source domain be represented by $(x^S , y^S ) = (x^S_k , y^S_k)_{k=1}^{m_S}$ , and the target domain be represented by $x^T = (x_k^T)_{k=1}^{m_T}$. The number of observations in the source and target domains are denoted by $m_S$ and $m_T$ respectively. The main challenge of domain adaptation is to develop a predictor that performs well in the target domain by leveraging the similarities between the two domains. One way to accomplish this is by making assumptions about how the joint distribution $P(X, Y)$ changes across the domains. In the case of \textit{covariate shift}, the marginal distribution $P(X)$ changes while the conditional distribution $P(Y|X)$ remains the same. However, in real-world scenarios, $P(Y|X)$ may also change, requiring further assumptions. One such assumption is that the joint distribution can be factored into $P(Y)$ and $P(X|Y)$, allowing changes in $P(Y)$ and $P(X|Y)$ to be addressed independently. The problem is then brokee tudied  a cial10.5555/3241691.324170810.5555/3241691.3241ution of features and labels.
\subsection{Motivation}

Unsupervised domain adaptation (UDA) is a technique used in machine learning where a model is trained on labeled data from a source domain that has similar characteristics to the target domain, but where the target domain lacks labeled data. The goal is to create a model that will perform well on the target domain despite not having labeled data from that domain. In UDA, the source and target domains are not directly related, so the model has to learn how to generalize across domains.\\

The first reason to be engaged in this field is a scarcity of data. It is known that collecting labeled data in the target domain can be expensive and time-consuming. UDA allows us to use the available labeled data in the associated source domain to learn representations that generalize well to the target domain without requiring additional labeled data. Minimizing the discrepancy between domains, the model can learn more robust and transferable representations, which leads us to the second reason -- improved generalization and domain robustness. The last reason is that UDA allows models to adapt to new environments. I reckon that this is a common situation in real applications, when models are trained on specially prepared data, and then applied to all other data types.

\newpage

\section{State-of-the-art}

\textit{In this section, we discuss main purposes, approaches and algorithms that specialists in the field of domain adaptation use in their research. In this section all figures are taken from the articles.}\\

\subsection{UDA by Backpropagation}

\begin{figure}[H]
    \centering
    \includegraphics[width=0.8\textwidth]{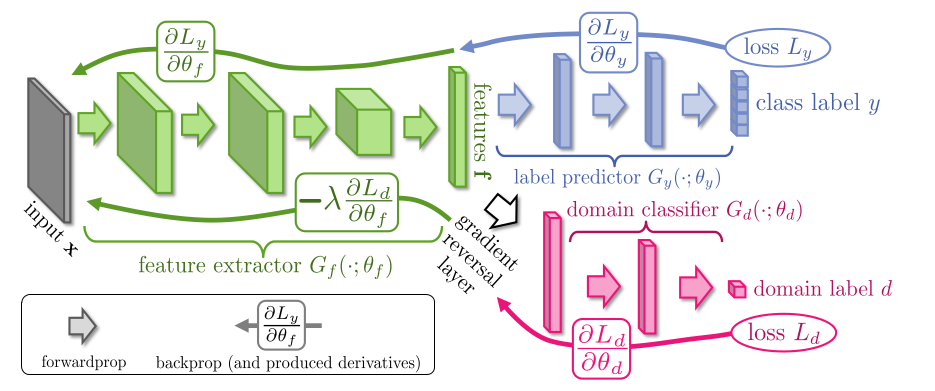}
    \caption{The proposed architecture includes a deep feature extractor (shown in green), a deep label predictor (shown in blue) and a domain classifier (shown in red). The domain classifier is connected to the feature extractor via a gradient reversal layer, which multiplies the incoming gradient by a negative constant during backpropagation-based training. The gradient reversal layer ensures that the feature distributions over the two domains become as similar as possible (i.e., indistinguishable by the domain classifier), resulting in domain-invariant features.}
    \label{fig: uda_back}
\end{figure}

The purpose of the article "Unsupervised Domain Adaptation by Backpropagation" written by Yaroslav Ganin and Victor Lempitsky \cite{ganin2015unsupervised} is to tackle the problem of domain shift in machine learning and to propose a solution to this problem using a neural-network model with few standard layers and gradient reversal layer (GRL). The GRL makes the network to learn domain-invariant features by minimizing the difference between the distributions of the source and target domains. The architecture of the model is shown below (see Figure \ref{fig: uda_back})

The authors introduce an architecture that predicts both the label $y \in Y$ and the domain label $d \in \{0, 1\}$ for each input $\textbf{x}$. The architecture consists of three parts: feature extractor $\bold{f} = G_f(\textbf{x}, \theta_f)$, where $\theta_f$ is a vector that represents the parameters of all its layers; label predictor $G_y$ that maps the features obtained after feature extractor to the label $y$, with $\theta_y$ representing its parameters; domain classifier $G_d$ maps the same feature vector $\bold{f}$ to the domain label $d$, with $\theta_d$ representing its parameters. The purpose to minimize the label prediction loss for the source domain and simultaneously make the features $\textbf{f}$ invariant to the domain. To achieve this, the authors optimize the parameters $\theta_f$ of the feature mapping to maximize the loss of the domain classifier, however the parameters $\theta_d$ are optimized to minimize the loss of the domain classifier. The authors consider the loss

\begin{equation}
\begin{split}
E(\theta_f, \theta_y, \theta_d) & = \sum_{\substack{i=1..N \\ d_i = 0}} L_y(G_y(G_f(\textbf{x}_i; \theta_f); \theta_y), y_i) - \lambda \sum_{i = 1..N} L_d(G_d(G_f(\textbf{x}_i; \theta_f); \theta_d), y_i) \\ & = \sum_{\substack{i=0..N \\ d_i = 0}} L_y^i(\theta_f, \theta_y) - \lambda \sum_{i=1..N} L_d^i(\theta_f, \theta_d)
\end{split}
\end{equation} 

 where $L_y$ and $L_d$ are label prediction and domain classification losses, respectively. (index $i$ means the i-th example). It is considered the parameters $\hat{\theta}_f, \hat{\theta}_y, \hat{\theta}_d$ to gain a saddle point

\begin{equation}
\begin{split}
(\hat{\theta}_f , \hat{\theta}_y ) = \arg \min_{\theta_f, \theta_y} E(\theta_f, \theta_y, \hat{\theta}_d)
\\
\hat{\theta}_d = \arg \max_{\theta_d} E(\hat{\theta}_f, \hat{\theta}_y, \theta_d)
\end{split}
\end{equation}

During learning, the trade-off between the two objectives that shape the features is controlled by the parameter $\lambda$. The following stochastic updates can find a saddle point

\begin{equation}
\begin{split}
\theta_f & \leftarrow \theta_f - \mu \left( \dfrac{\partial L_y^i}{\partial \theta_f} - \lambda \dfrac{\partial L_d^i}{\partial \theta_f} \right) \\ 
\theta_y & \leftarrow \theta_y - \mu \dfrac{\partial L_y^i}{\partial \theta_y} \\ 
\theta_d & \leftarrow \theta_d - \mu \dfrac{\partial L_d^i}{\partial \theta_d} 
\end{split}
\end{equation} 

where $\mu$ is a learning rate. These updates are similar to SGD but with a $-\lambda$ factor in the first update to prevent dissimilar features across domains. Therefore, the authors introduce a GRL that acts as an identity transform during forward propagation but multiplies the gradient by $-\lambda$ during backpropagation.
\subsection{Semantic Representations for UDA}
Next, we continue with the article \textit{"Learning Semantic Representations for Unsupervised Domain Adaptation"} written by Xie, Shaoan, et al. \cite{xie2018learning} The main purpose of the article is to propose a new method for unsupervised domain adaptation that utilizes semantic information to learn domain-invariant representations. The authors propose a domain adaptation algorithm which is based on the idea of using an adversarial learning to learn a feature representation that is invariant to domain shifts. 

\begin{figure}[H]
    \centering
    \includegraphics[width=0.7\textwidth]{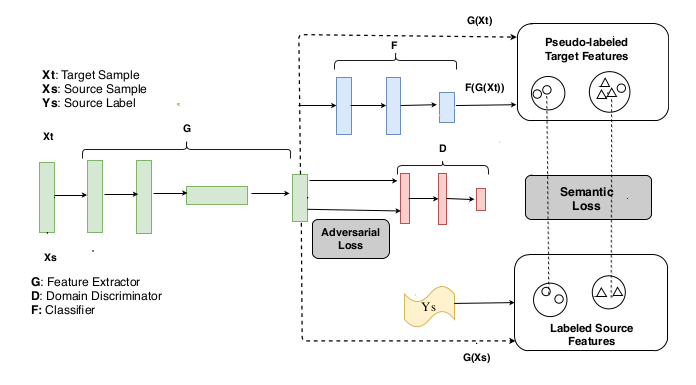}
    \caption{The authors use standard source classification loss with the domain adversarial loss to align distribution for two domains. It is showed that the performance of the domain adaptation method improved significantly on several benchmark datasets by aligning the centroids. Global centroids $C_S^k$ and $C_T^k$ is maintained for each class at feature level.}
    \label{fig: sem_rep}
\end{figure}

 The authors train a feature extractor and then use it to map the input data to a high-dimensional feature space, and a domain classifier that predicts the domain label of the input data (see Figure \ref{fig: sem_rep}). The feature extractor \textbf{G} is trained to confuse the domain classifier \textbf{D}, while the domain classifier is trained to correctly predict the domain label. In this way, the feature extractor is encouraged to learn features that are invariant to domain shifts, while still being discriminative for the task.\\

 First, the authors denote the cross entropy loss for the source domain as $L_C(X_S, Y_S)$. Then, the discrepancy between source domain and target domain is supposed to be 
\begin{equation}
 L_{DC}(X_S, X_T) = d(X_S, X_T) = \mathbb{E}_{x \sim D_S} [\log(1 - D \circ G(x))] + \mathbb{E}_{x \sim D_T} [\log(D \circ G(x))] 
\end{equation} 

Moreover, the authors introduce one more loss, which targets the semantic representation. Centroid alignment is used for this purpose. By computing the centroid for each class, both correct and incorrect pseudo-labeled samples are utilized together: 

\begin{equation}
L_{SM}(X_S, Y_S, X_T) = \sum_{k=1}^K \Phi(C_S^k, C_T^k)
\end{equation}
where $C_S^k, C_T^k$ are centroids for each class and $\Phi(x, x') = \| x - x' \|^2$. This approach aims to cancel out the negative effects caused by inaccurate pseudo labels with accurate ones. Thus, the authors get the following total loss

 \begin{equation}
 L(X_S, Y_S, X_T) = L_C(X_S, Y_S) + \lambda L_{DC}(X_S, X_T) + \gamma L_{SM}(X_S, Y_S, X_T)
 \end{equation}
 where $\lambda$ and $\gamma$ are responsible for the balance between the classification
loss, domain confusion loss and semantic loss. In the article, algorithm of \textbf{moving average centroid alignment} is presented that allows to align the centroids in same class
but different domains to achieve semantic transfer for UDA.
\subsection{Fixbi for UDA}

\begin{figure}[H]
    \centering
    \includegraphics[width=\textwidth]{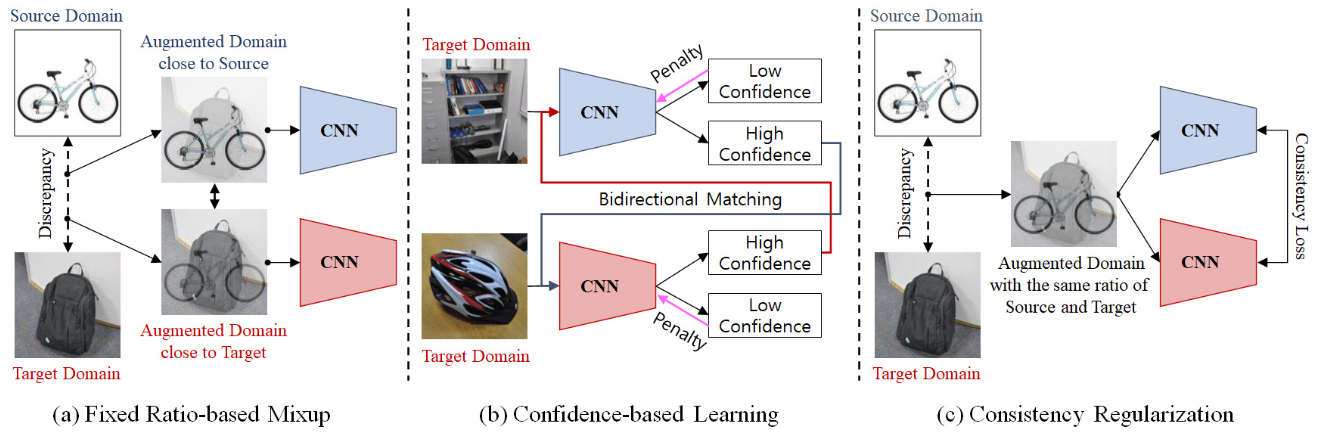}
    \caption{All parts of FixBi training model: (a) fixed ratio-based mixup, (b)
confidence-based learning and (c) consistency regularization.}
    \label{fig: fixbi}
\end{figure}

The purpose of the article "Fixbi: Bridging domain spaces for unsupervised domain adaptation" written by Jaemin Na, Heechul Jung et al. \cite{na2021fixbi} is to propose a  fixed ratio-based mixup method to address the problem of large domain discrepancies. The authors mix up images and then fed them into neural networks to achieve greater reliability in learning from corrupted labels. It is proposed to use two predetermined mixup ratios $\lambda_sd$ and $\lambda_td$ for the source and target domain respectively. 
Denote input samples and their labels for source and target domain as $(x_i^s, y_i^s)$ and $(x_i^t, \hat{y}_i^t)$, the authors define mixup configurations in the following way:

\begin{equation}
\begin{split}
 \tilde{x}^{st}_i &= \lambda x_i^s + (1 - \lambda)x_i^t\\
 \tilde{y}^{st}_i &= \lambda y_i^s + (1 - \lambda)\hat{y}_i^t,
\end{split}
\end{equation} 

where $\lambda \in \{\lambda_{sd}, \lambda_{td} \}$ and $\lambda_{sd} + \lambda_{td} = 1$, $\hat{y}_i^t$ is the pseudo-labels for the target samples. By leveraging the fixed ratio-based mixup, it is constructed two neural networks with different perspectives: the "source-dominant model" (SDM) and the "target-dominant model" (TDM) (see Figure \ref{fig: fixbi}).

The SDM provides robust supervision for the source domain but relatively weak supervision for the target domain, while the TDM has strong supervision for the target domain but weaker supervision for the source domain. Thus, denoting $p(y| \tilde{x}^{st}_i)$ as a predicted class distribution, it is defined fixed ratio-based mixup function 

\begin{equation}
\label{eq:fixbi1}
L_{fm} = \dfrac{1}{B} \sum_{i = 1}^B \hat{y}^{st}_i \log (p(y|\tilde{x}^{st}_i)),
\end{equation}

where $\hat{y}^{st}_i = \arg \max p(y|\tilde{x}^{st}_i)$ and $B$ is the size of a mini-batch. In order to have connections between source and target domains, it is suggested to use a confidence-based learning approach whereby one model educates the other using positive pseudo-labels, or penalties itself using negative pseudo-labels. Positive pseudo-labels means labels which predictions are above a specific threshold, then the authors use them in training the second model by utilizing a conventional cross-entropy loss. Thus, denote p and q as distributions of two models, the authors get the following loss function

\begin{equation}
L_{bim} = \dfrac{1}{B}\sum_{i=1}^B \mathds{1}(\max (p(y|x_i^t) > \tau)\hat{y}_i^t \log (q(y|x_i^t)),
\end{equation}

where $\hat{y}_i^t = \arg \max p(y|x_i^t)$. In contrast, a negative pseudo-label refers to the top-1 label predicted by the network with a confidence below the threshold $\tau$. The function of self-penalization is defined as follows:

\begin{equation}
L_{sp} = \dfrac{1}{B}\sum_{i=1}^B \mathds{1}(\max (p(y|x_i^t) < \tau)\hat{y}_i^t \log (1 - p(y|x_i^t))
\end{equation}
Furthermore, the threshold is changed adaptively during training. In addition, it is introduced the following expression: 

\begin{equation}
\label{eq:fixbi4}
L_{cr} = \dfrac{1}{B} \sum_{i=1}^B \| p(y|\tilde{x}^{st}_i) - q(y|\tilde{x}^{st}_i)\|^2_2
\end{equation}
that represents consistency regularization to guarantee a stable convergence during the training of both models.
\subsection{Spherical Space DA with Pseudo-label Loss}

Now we are going to the next article \textit{"Spherical Space Domain Adaptation with Robust Pseudo-label Loss"} written by  Xiang Gu, Jian Sun, and Zongben Xu. \cite{gu2020spherical} The authors propose a spherical space representation of data, which allows them to get more effective feature extraction and better adaptation across domains. One approach associated with increasing performance with differences in data distribution between source and target domain is to use pseudo-labels. However, the use of pseudo-labels can be problematic in the presence of noisy or incorrect labels. To tackle this problem, the authors map the data to a high-dimensional sphere and introduce a new loss function, called the robust pseudo-label loss, which is designed to address the problem of noisy or incorrect labels in the target domain. 

\begin{figure}[H]
    \centering
    \includegraphics[width=\textwidth]{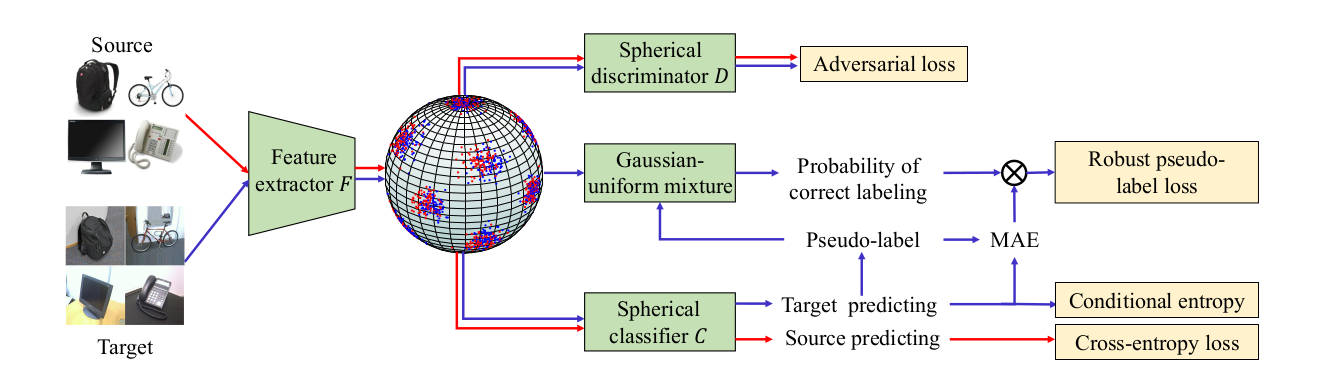}
    \caption{Robust Spherical Domain Adaptation (RSDA) method proposed by the authors consists of a feature extractor F, which is a deep convolutional network, used to extract features that are then embedded onto a sphere, a spherical classifier and a discriminator that predict class labels and domain labels, respectively. Target pseudo-labels and features passing through the Gaussian-uniform mixture model are used to estimate the posterior probability of correct labeling.}
    \label{fig: RSDA_scheme}
\end{figure}

\begin{figure}[H]
    \centering
    \includegraphics[width=0.7\textwidth]{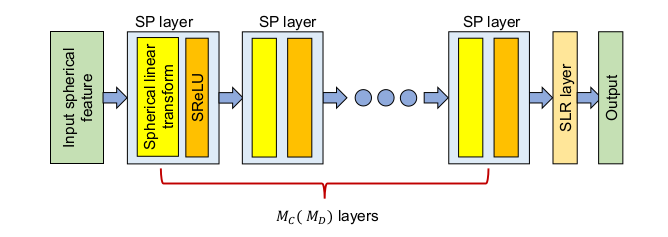}
    \caption{Spherical neural network structure}
    \label{fig: sphere_layers}
\end{figure}

In the Figure \ref{fig: RSDA_scheme} we can see spherical domain adaptation method. Domain invariant features are learned by adversarial training, entirely in the spherical feature space.  The feature extractor F is utilized to normalize the features to map onto a sphere. The classifier C and discriminator D are defined in the spherical feature space, consisting of spherical perceptron layers and a spherical logistic regression layer (see Figure \ref{fig: sphere_layers}). 

Although the use of a spherical element reduces feature dimension by one, it simplifies the domain adaptation process by eliminating differences in norms. The authors define spherical adversarial training loss as follows:

\begin{equation}
    L = L_{bas}(F, C, D) + L_{rob}(F, C, \phi) + \gamma L_{ent}(F)
\end{equation}

This lost consists of three parts: basic loss, robust pseudo-label loss and conditional entropy loss. Let's start with the first one. To align features, the authors utilize basic loss which is defined as an adversarial domain adaptation loss:
\begin{equation}
L_{bas}(F, C, D) = L_{src}(F, C) + \lambda L_{adv}(F, D) + \lambda' L_{sm}(F),
\end{equation}
where $L_{src}$ is a cross-entropy loss for the source domain, $L_{adv}$ is an adversarial training loss and $L_{sm}$ is a semantic loss. The second one is conditional entropy loss which is used to keep the learned features away from the classification boundary:
\begin{equation}
L_{ent}(F) = \dfrac{1}{N_t} \sum_{j=1}^{N_t} H(C(F(x_j^t))),
\end{equation}
where $H(\cdot)$ denotes the entropy of a distribution. Additionally, the authors propose robust pseudo-label loss to increase robustness of the model. Denote $\tilde{y}_j^t = \arg \max_k [C(F(x_i^s))]_k$ as a pseudo-lebel of $x_j^t$ where $[\cdot]_k$ means the $k$-th element. To be ensured in precision of pseudo-labels, it is assumed to use new random variable $z_j \in \{0, 1\}$ for each pair $(x_j^t, \tilde{y}_j^t)$ that specify the correctness of the data (1 is correct, 0 is not). Let the probability of correct labeling be $P_\phi (z_j = 1| x_j^t, \tilde{y}_j^t)$ and $\phi$ to its parameters, then robust loss is defined as follows:

\begin{equation}
L_{rob}(F, C, \phi) = \dfrac{1}{N_0}\sum_{j=1}^{N_t} w_\phi(x_j^t) \mathcal{J}(C(F(x_j^t)), \tilde{y}_j^t),
\end{equation}
where $N_0 = \sum_{j=1}^{N_t} w_\phi(x_j^t)$ and $\mathcal{J}(\cdot, \cdot)$ is mean absolute error (MAE). The function $w_\phi (x_j^t)$ is defined using the posterior probability of correct labeling 

\begin{equation}
w_\phi(x_j^t) = \left\{\begin{split}
\gamma_j,& \;\; \text{if } \gamma_j \geq 0.5,\\
0,& \;\; \text{otherwise},
\end{split}\right.
\end{equation}
where $\gamma_j = P_\phi(z_j = 1| x_j^t, \tilde{y}_j^t)$. By utilizing a Gaussian-uniform mixture model in spherical space based on pseudo-labels, the authors model the probability $P_\phi(z_j = 1| x_j^t, \tilde{y}_j^t)$ as a function of the feature distance between the data and the center of the corresponding class. Thus, samples from target domain with a probability of correct labeling below 0.5 can be discarded. For further details of computing posterior probability, please refer to the article \cite{gu2020spherical}.  
\subsection{DA with Invariant Representation Learning}
 Next, we move on to the article \textit{"Domain Adaptation with Invariant Representation Learning: What Transformations to Learn?"} written by Stojanov, Petar, et al. \cite{stojanov2021domain} The researchers focus on the conditional shift scenario, where the data-generating process is utilized to (\textbf{i}) explain why two distinct encoding functions are required to infer the latent representation, (\textbf{ii}) improve an implementation of these functions, and (\textbf{iii}) impose meaningful structure on the latent representation $Z$ to increase prediction accuracy in the target domain.

\begin{figure}[H]
    \centering
    \includegraphics[width=0.5\textwidth]{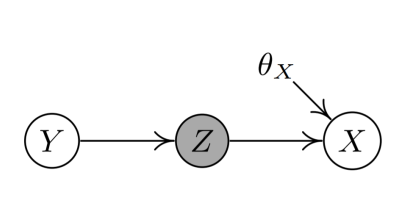}
    \caption{Data-generating process under conditional shift}
    \label{fig: DA_transf}
\end{figure}

Let's consider the data-generating process shown in the Figure \ref{fig: DA_transf} to understand what information is required for learning. The label $Y$ is generated first from its prior distribution $P(Y)$. Then, the invariant representation $Z$ is generated from $Y$ through $P(Z|Y)$, and $X$ is generated from $P(X|Z; \theta_X)$, where $\theta_X$ represents the changing parameters of $P(X|Y)$ across domains. We can consider $Z$ as a latent representation of our data. The variable $\theta_X$ may correspond to environment-specific changes that are irrelevant for predicting the class $Y$. Generally speaking, $Z$ is conditionally dependent on $\theta_X$ given $X$, although they may be marginally independent. Therefore, to recover $Z$ given $X$, the information of $\theta_X$ should also be considered in the transformation (see detailed in the article to understand clearly how authors measure the influence of $\theta_X$). The authors made two key observation associated with the data-generating process. Firstly, the encoder function $\phi$ requires $\theta_X$ as an input in addition to X. Secondly, assuming that $\theta_X$ has minimal influence on the relationship between $X$ and $Z$, allowing us to use a single encoder $\phi(X, \theta_X)$ instead of two separate encoders. A decoder function $\widetilde{\phi}$ that restricts the influence of $\theta_X$, acting as a regularizer on the encoder $\phi$, in order to retain important semantic information.

\begin{figure}[H]
    \centering
    \includegraphics[width=0.85\textwidth]{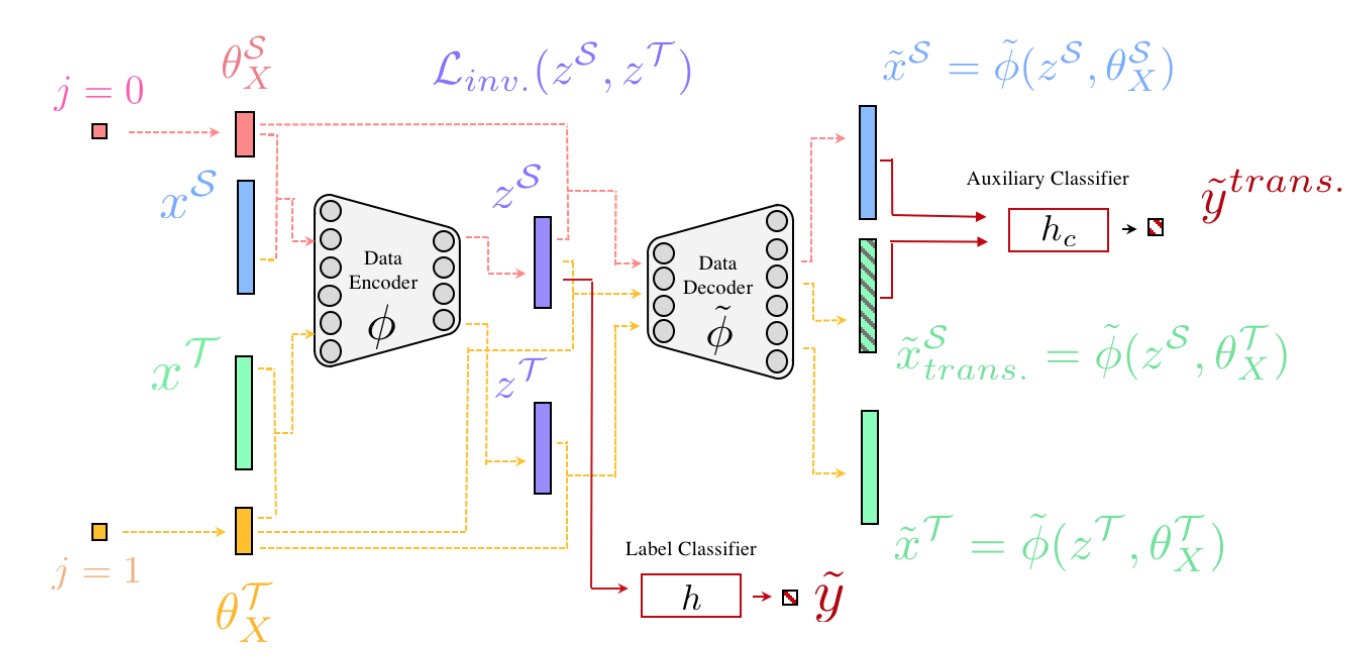}
    \caption{Autoencoder framework used in this article}
    \label{fig: DA_cnn}
\end{figure}

Thus, the authors proposed a domain-adaptation network, which is shown in the Figure \ref{fig: DA_cnn}, where $\theta_X \in \{\theta_X^S, \theta_X^T\}$ parameters for source and target domains respectively. 
\subsection{Domain Adaptation for Segmentation with CBST}

\begin{figure}[H]
    \centering
    \includegraphics[width=0.8\textwidth]{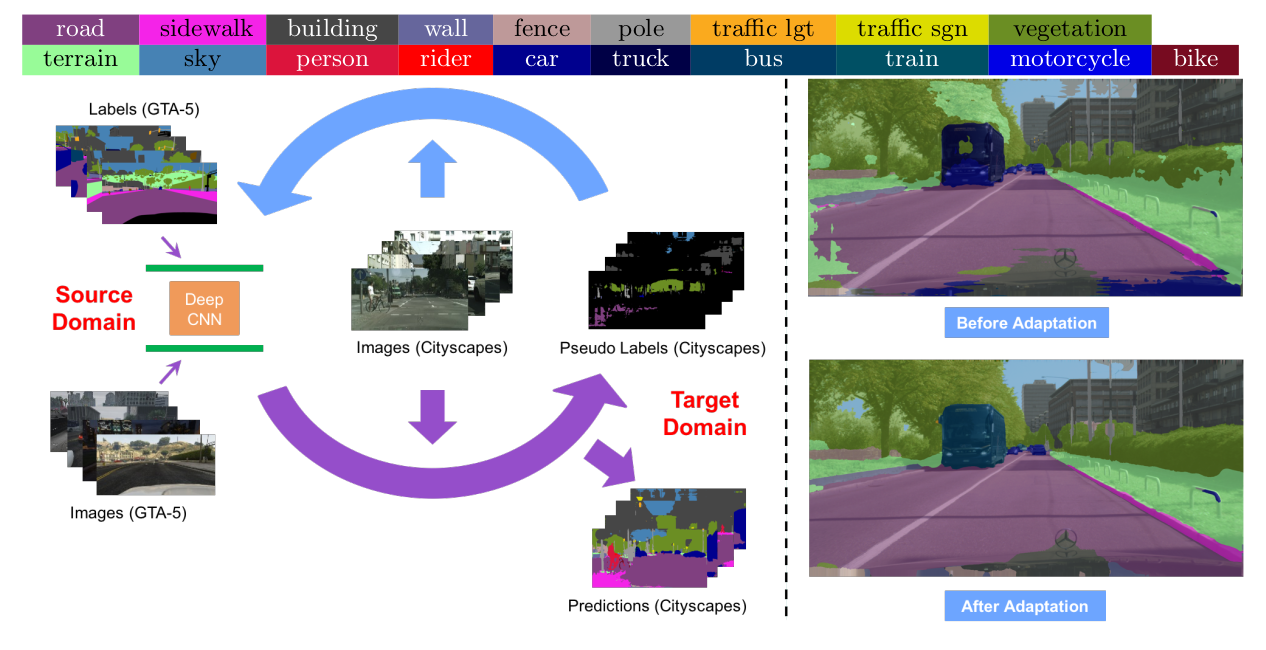}
    \caption{ On the left side, the self-training framework for UDA is presented. On the right side, obtained results before and after adaptation for the Cityscapes dataset.}
    \label{fig: sem_seg}
\end{figure}

The main purpose of the paper \textit{"Domain Adaptation for Semantic Segmentation via Class-Balanced Self-Training"} written by Zou, Yang, et al. \cite{zou2018unsupervised} propose a new UDA framework for semantic segmentation based on iterative self-training procedure. A novel technique, referred to as Class-Balanced Self-Training (CBST), has been suggested by the authors, which aims to adapt the segmentation model from the source domain to the target domain by leveraging unlabeled target data. In the Figure \ref{fig: sem_seg}, the authors present a structure and the results of their deep self-training framework using two datasets: GTA 5 \cite{richter2016playing} and Cityscapes \cite{cordts2016cityscapes}.

The CBST approach is based on two main components: a class-balancing strategy and a self-training algorithm. The class-balancing strategy aims to address the problem of class imbalance between the source and target domains, which can negatively impact the performance of the segmentation model. The authors change the loss function using parameters that determine the proportion of selected pseudo-labels due to balance the class distribution during the self-training process. Furthermore, when the images in the source and target domains are similar, spatial prior knowledge can be effectively utilized to adapt models. For this purpose, the authors count the class frequencies in the source domain using Gaussian kernel. The experimental results show that the CBST approach outperforms several state-of-the-art unsupervised domain adaptation methods for semantic segmentation. \newpage

\section{Contribution}
The "Contribution" section of the thesis highlights the developed new method called \textbf{DannFixbi}, which combines the Fixbi approach and the backpropagation approach. In an attempt to implement the state-of-the-art method, extensive research has been conducted, and existing approaches have been implemented. The decision to combine these two approaches was based on their respective strengths and the potential for mutually beneficial interaction between them. By incorporating the Fixbi technique, which addresses domain shift, and leveraging the benefits of backpropagation, \textbf{DannFixbi} aims to enhance the performance and robustness of domain adaptation in the field of images. The development of \textbf{DannFixbi} represents an original contribution to the field. \\

This new method consists of two neural networks, which are trained using a modified version of the Fixbi approach. To enhance the performance of this method, two domain classifiers are added to each of these two networks. Two different approaches have been explored for incorporating these domain classifiers.\\

\begin{figure}[H]
    \centering
    \includegraphics[width=0.55\textwidth]{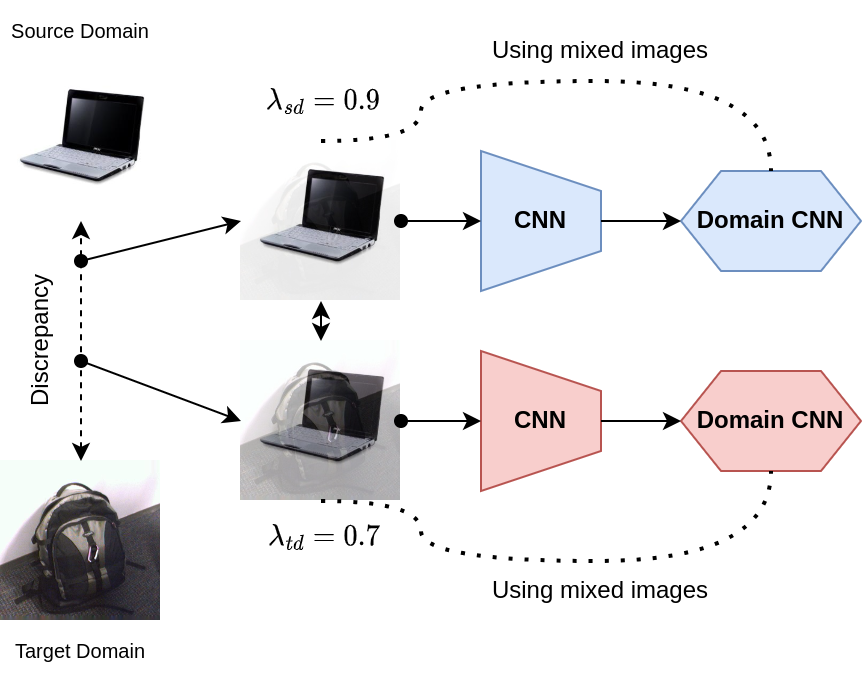}
    \caption{\textbf{DannFixbi} architecture of the first approach.}
    \label{fig:dannfixmix}
\end{figure}

The first approach involves adding a domain classifier to each neural network (see Figure \ref{fig:dannfixmix}). During training, images obtained by mixing from the source and target domains with predefined mixup ratios are fed into these classifiers. 

For mixed images, the following loss functions is used:

\begin{equation}
    L_{dom} = \alpha L_{ds}(\hat{X}, Y_{s}) + (1 - \alpha) L_{dt}(\hat{X}, Y_{t}) ,
\end{equation}

where $\hat{X}$ is mixed images, $Y_{s}$ is source domain labels, $Y_{t}$ is target domain labels and $\alpha \in \{\lambda_{sd}, \lambda_{td}\}$. $L_{ds}$ and $L_{dt}$ presents cross entropy loss for source and target domains, respectively.\\

\begin{figure}[H]
    \centering
    \includegraphics[width=0.55\textwidth]{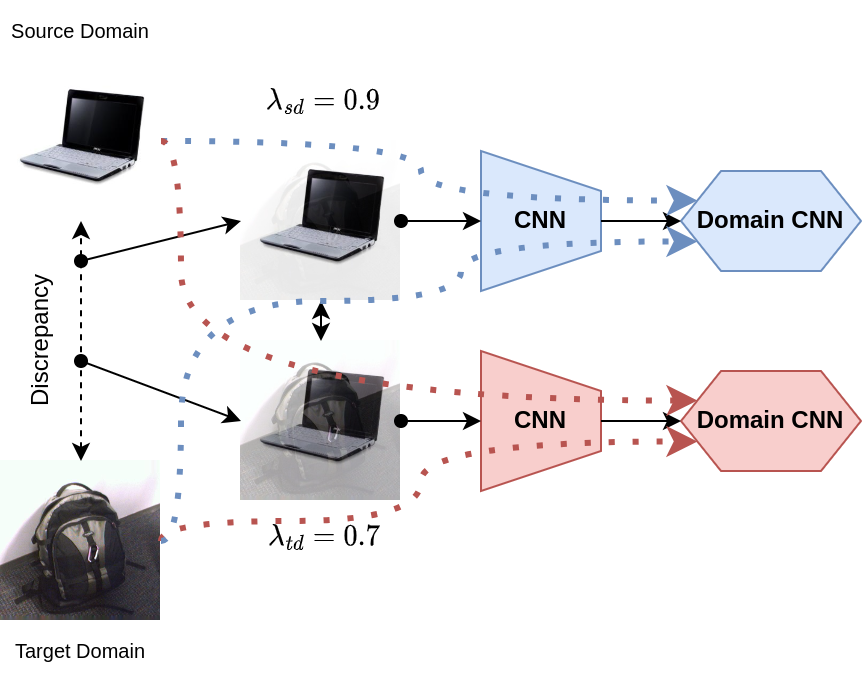}
    \caption{\textbf{DannFixbi} architecture of the second approach.}
    \label{fig:dannfixsep}
\end{figure}

The second approach is to use a domain classifier for each net with images from the source and target domains without any mixing, similar to the backpropagation method (see Figure \ref{fig:dannfixsep}).\\

The second approach utilizes the following loss function for domain classification:

\begin{equation}
L_{dom} = L_{d}(X_{st}, Y_{st}) ,
\end{equation}

where $X_{st}$,  $Y_{st}$ denotes source and target images and domain labels, respectively, and $L_{d}$ is a cross entropy loss. \\

The total loss for the new method is calculated as the sum of the loss from the Fixbi method and the domain loss described earlier:

\begin{equation}
L_{total} = \beta L_{fixbi} + \gamma L_{dom}
\end{equation}

Here, $\beta$ and $\gamma$ represent the weights assigned to the Fixbi loss and the domain loss, respectively. The values of these weights determine the relative importance of each component in the overall loss calculation. Further, unless otherwise stated, the values of alpha and beta are assumed to be equal to 1. The Fixbi loss, denoted as $L_{fixbi}$, is composed of several summands described in the  equations \ref{eq:fixbi1} -- \ref{eq:fixbi4}: 

\begin{equation}
L_{fixbi} = L_{fm} + L_{sp} + \mathds{1}\{e > k\} \left(L_{bim} + L_{cr}\right)
\end{equation}
where $e$ denotes a current epoch, and $k$ is  warm-up epochs. To establish independent characteristics for the two networks, it is introduced a warm-up period of $k$ epochs. During this phase, each network is trained separately using the fixed ratio-based mixup and self-penalization techniques. Once an enough amount of training has been completed, bidirectional matching loss is added, which helps networks train collaboratively, exchanging knowledge and benefiting from each other's insights. \\

The new method, referred to as \textbf{DannFixbi}, demonstrates improved performance and robustness in unsupervised domain adaptation for image analysis. This contribution represents a unique approach to UDA, offering valuable insights and potential for future developments in this field. All the results obtained are presented in the "Experiments" section.

\newpage
\section{Experimental Setup} \label{section: copy me}

In this part, we start with description of different datasets that are commonly used in transfer learning. Then, we will continue with implementation details and experiments that have been conducted. Code is available at \href{https://github.com/Jetwev/domain-adaptation}{https://github.com/Jetwev/domain-adaptation}.

\subsection{Datasets}

The most popular datasets are \textbf{Office-31}, \textbf{ImageCLEF-DA}, \textbf{Office-Home}, \textbf{DomainNet} and \textbf{VisDA-2017}. Detailed discussion of each of them is given below:

\begin{itemize}
    \item \textbf{Office-31} \cite{saenko2010adapting} consists of 4,110 images categorized into 31 classes, which are distributed across three separate domains: Amazon (A), Webcam (W), and Dslr (D).
    \item \textbf{ImageCLEF-DA}, utilized in \cite{long2017deep}, includes three distinct domains: Caltech-256 (C), ImageNet ILSVRC 2012 (I), and Pascal VOC 2012 (P). There are 600 images in each domain and 50 images for each category.
    \item \textbf{Office-Home} \cite{venkateswara2017deep}, includes four absolutely different domains: Artistic images (Ar), Clip Art (Cl), Product images (Pr) and Real-World images (Rw). This dataset contains 15~500 images in 65 object classes, which makes it more complex than Office-31.
    \item \textbf{VisDA-2017} \cite{peng2017visda} consists of 12 classes shared between two very different domains: Synthetic and Real. It contains synthetic images (training set) and real-world images (test set). The dataset was designed to have a large domain gap, which makes it a challenging benchmark for domain adaptation methods.
    \item \textbf{DomainNet} \cite{peng2019moment} is a large-scale visual recognition dataset designed to evaluate domain adaptation algorithms, which consists of almost 600 thousand images and includes 345 classes.
\end{itemize}

\newpage

\subsection{Implementation details}

At the beginning, it was necessary to start with some approaches to check their performance and have a possibility to compare results. Four different methods described in the papers have been chosen for the study:

\begin{itemize}
    \item \textbf{Source only} is a method where a model is trained solely on the source domain data without any adaptation to the target domain.
    \item \textbf{Domain-Adversarial Neural Network (Dann)} is domain adaptation technique that aims to learn a domain-invariant feature representation by aligning the feature distributions of the source and target domains. The architecture of Dann consists of three components: a feature extractor network, a label predictor network, and a domain classifier network, and is described in more details in section 2.1.
    \item \textbf{Moving Semantic Transfer Network (Mstn)} The key idea behind Mstn is to add semantic transfer loss to the Dann approach. In the section 2.2, it is proposed to use average centroid alignment for aligning the feature distributions of the source and target domains. The architecture is the same as in the Dann method. 
    \item  \textbf{Fixbi} is the approach described in details in the section 2.3. The main idea is to train two neural networks, allowing models to learn from each other or on their own results. For this purpose, the authors add bidirectional matching and self-penalization losses.
\end{itemize}

\textbf{CNN architectures.} For all approaches, pretrained Resnet50 \cite{he2016deep} is utilized as the backbone network. The weights for the neural network can be downloaded from this \href{https://download.pytorch.org/models/resnet50-19c8e357.pth}{link}. Resnet50 has been pretrained on large image datasets such as ImageNet, which means that the network has already learned to recognize a wide range of features in images. Resnet50 is a convolutional neural network architecture consisting of 50 layers. This is a variant of the Resnet family of neural networks, which are designed to solve the vanishing gradient problem in deep neural networks. Resnet networks achieve this by using short connections between layers, which allow gradients to move more easily during backpropagation. Resnet50 is a widely used architecture in many articles, which makes it a good choice for research.\\ 

Resnet50 is used as a \textit{Feature extractor} in all considering methods. \textit{Label predictor} is a simple network that consists of two fully connected layers ($2048 \rightarrow 256 \rightarrow \textit{number of classes}$). \textit{Domain classifier} architecture represents several fully connected layers with a ReLU activation function and dropouts between each two fully connected layers. Using dropouts can reduce the sensitivity of the model to specific features in the input data and encourage the model to learn more generalizable features. This can lead to better performance on new, unseen data and can prevent overfitting. \\


\textbf{Learning rate schedulers.} Learning rate is an important hyperparameter that determines the step size at which the optimizer updates the model's parameters during training. There are many of them and it can be challenging to find the optimal learning rate, as setting it too high can cause the model to diverge, while setting it too low can slow down the learning process. Thus, in this study it is utilized two different learning rate schedulers: \textit{CustomLRScheduler} and \textit{CosineAnnealingLR}. The implementation of the first one follows the rules that are described in \cite{ganin2015unsupervised} 

\begin{equation}
    \eta_p = \dfrac{\eta_0}{(1 + \alpha \cdot p)^\beta},
\end{equation}

where $p$ linearly increases from $0$ to $1$, and the values $\eta_0$, $\alpha$, and $\beta$ are set to $0.01$, $10$, and $0.75$, respectively. The second one, \textit{CosineAnnealingLR} is a popular learning rate scheduler utilized in deep learning. It systematically reduces the learning rate over multiple epochs in a cyclical manner. Initially, the learning rate starts at its maximum value and then gradually decreases until it reaches the minimum value. Upon reaching the minimum value, the cycle restarts, and the learning rate returns to its maximum value. This process continues until the end of the training, which is usually determined by the total number of epochs or a predefined stop criterion. By starting with a higher learning rate and gradually decreasing it, the model can avoid getting stuck in local minima and converge to a better global minimum. The formula for the \textit{CosineAnnealingLR} scheduler is:

\begin{equation}
\eta_t = \eta_{min} + \dfrac{1}{2}(\eta_{max} - \eta_{min}) \left(1 + \cos \left( \dfrac{T_{cur}}{T_{max}} \pi \right)\right),
\end{equation}

where $\eta_{max}$ is your initial learning rate, $\eta_{min}$ -- minimum learning rate value, $T_{cur}$ is the number of epochs since the last start, $T_{max}$ -- the total number of epochs. More detailed information about \textit{CosineAnnealingLR} can be found \href{https://pytorch.org/docs/stable/generated/torch.optim.lr_scheduler.CosineAnnealingLR.html}{here}.\\

\textbf{Optimizers.} This study uses two popular optimization algorithms - stochastic gradient descent (\textit{SGD}) and adaptive moment estimation (\textit{Adam}). Both algorithms are commonly employed in deep learning to optimize the parameters of a neural network and improve its performance. \\

\textit{SGD} is a simple and popular optimization algorithm that updates the weights of a model in the direction of the negative gradient of the loss function. One limitation of \textit{SGD} is that it can get stuck in local minima and struggle with noisy or sparse gradients. To tackle this problem, several modifications can be used. In PyTorch, the \textit{SGD} optimizer has several hyperparameters that can be tuned to improve its performance. The following parameters (except learning rate) are considered in this study: 

\begin{itemize}
    \item \textbf{Momentum} is a hyperparameter that determines how much past gradients affect the current gradient update. It helps to minimize the impact of the noise and fluctuations in the gradient updates. However, setting the momentum too high can also lead to slower convergence.
    
    \item  \textbf{Weight decay} is a form of L2 regularization that adds a penalty term to the loss function during training. This penalty term is proportional to the square of the weights in the network, which encourages the model to use smaller weights and reduce overfitting.

    \item \textbf{Nesterov momentum} is a variant of momentum that takes into account the momentum term in the calculation of the gradient. This can help to reduce oscillations and improve convergence rates, especially in high-dimensional optimization problems.
\end{itemize}

\textit{Adam} is another optimization algorithm that is commonly used in deep learning. It is an extension of SGD. The key idea behind Adam is to maintain a separate adaptive learning rate for each parameter in the network, based on estimates of the first and second moments of the gradients. This makes Adam more effective than SGD for optimization problems with noisy or sparse gradients. However, it may not always be the best choice for every task and model architecture, so it's important to experiment with different optimization algorithms and settings to find the best approach for your specific problem. \textit{Adam} is considered in this study with default parameters, more information about the implementation and usage can be found at this \href{https://pytorch.org/docs/stable/generated/torch.optim.Adam.html}{link}.\\

\textbf{Pytorch Lightning.} PyTorch Lightning is a lightweight PyTorch wrapper that allows users to focus on the high-level design of their experiments and models, instead of dealing with the low-level implementation details. It provides a structured way to organize PyTorch code, making it easier to read and maintain.\\

PyTorch Lightning offers a range of benefits that make it a good choice for deep learning researches. Firstly, it offers a modular design that makes it easy to organize code. It gives you a convenient and user-friendly interface to manage and run experiments. Moreover, all these benefits can help to improve your productivity. Secondly, PyTorch Lightning makes it easier to scale models to multiple GPUs, which can significantly reduce training times for large models. Finally, it is flexible and can be easily integrated with other PyTorch libraries. Overall, PyTorch Lightning is an excellent choice for researchers who want to focus on the research aspect of deep learning and leave the engineering components to the library. More information can be found on the official \href{https://www.pytorchlightning.ai/index.html}{website}.\\

\textbf{Weights and Biases.} Weights and Biases (WandB) is a platform  that provides a suite of tools to help developers and data scientists track and visualize their machine learning experiments. WandB makes it easy to log, keep track of your progress and compare different experiments, visualize model performance, and collaborate with team members. One of the main advantages of WandB is its integration with popular machine learning frameworks such as TensorFlow, PyTorch, and Keras. This means that you can easily log and track your model's hyperparameters and performance metrics during training and evaluation. Moreover, WandB is a cloud-based platform, which means that users can access their experiments and data from anywhere with an internet connection and also share them with colleagues and co-workers. For more detailed information, it is recommended to visit the official \href{https://wandb.ai/site}{website}.\\ 

\textbf{Batch size.} Different domains in your dataset can contain different number of images that makes your training process more complicated. To tackle this problem, it is proposed two approaches. \\

The first one is to find the ratio of the smaller dataset size to the larger one and concatenate the smaller dataset multiple times to ensure that the number of batches is aligned during the training loop. However, it is important to emphasize that with this approach, overfitting can occur if the appropriate number of epochs is not established. This is because the smaller dataset will be fed into the model more times than the larger one (depends on the ratio).\\

The second approach involves varying the number of images taken per batch for each domain. Applying this approach, it becomes possible to avoid concatenating the smaller dataset multiple times, which effectively reduces the amount of memory consumed. It is crucial to carefully consider the number of images per batch, as choosing a value that is either too high or too low can have negative consequences.\\

\textbf{Augmentation techniques.} Augmentation techniques for images are used to create variations and increase the size of the training dataset by applying a series of transformations. These techniques are widely employed in computer vision tasks, including image classification, object detection, semantic segmentation, etc. Augmentation helps increase the diversity of the dataset, leading to improved model generalization and robustness. PyTorch provides a variety of image augmentation techniques. The following transformations are used in this research:

\begin{itemize}
    \item \textbf{Normalize} -- normalizes the image by subtracting the mean value and dividing by the standard deviation. 
    \item \textbf{Resize} is a function that allows you to resize an image to a specific size.
    \item \textbf{RandomCrop} is a function that randomly crops a portion of the image.
    \item \textbf{CenterCrop} is a transformation that allows you to perform a center crop on an image.
    \item \textbf{RandomHorizontalFlip} -- randomly flips the image horizontally with a specified probability.
    \item \textbf{RandomVerticalFlip} - randomly flips the image vertically with a specified probability.
    \item \textbf{RandomRotation} is a function that randomly rotates the image by a given angle.
    \item \textbf{ColorJitter} is a transformation that allows you to adjust the brightness, contrast, saturation, and hue of the image.
    \item \textbf{ToTensor} is a specific function in PyTorch that is used to convert an image into a tensor.
\end{itemize}

These augmentation techniques can be applied individually or combined sequentially using the \textit{transforms.Compose} function. More detailed information about transformations and their usage can be found \href{https://pytorch.org/vision/stable/transforms.html}{here}. It's important to emphasize that the choice and combination of augmentation techniques depend on the specific task and dataset characteristics, and careful selection of them are crucial to achieve optimal results.\\

\textbf{OmegaConf.} OmegaConf is a library for Python that provides a convenient and flexible way to manage complex configurations in machine learning projects. It is designed to provides a number of features that can help to simplify the configuration process. Here are some reasons why OmegaConf can be a good choice:

\begin{itemize}
    \item It is easy to use and allows developers to define nested configurations and easily access and modify configuration values.
    \item OmegaConf supports a wide range of configuration formats, including YAML and JSON. This makes it flexible and easy to integrate in your project.
    \item It supports type checking, which can help to catch configuration errors and improve code quality.
\end{itemize}
 
To sum up, OmegaConf can be a good choice for Python developers who work on large and complex projects and want a flexible and powerful configuration system for their applications. Additional details can be found \href{https://omegaconf.readthedocs.io/en/2.3_branch/}{here}.

\newpage

\subsection{Experiments}

\begin{figure}[H]
    \centering
    \includegraphics[width=0.9\textwidth]{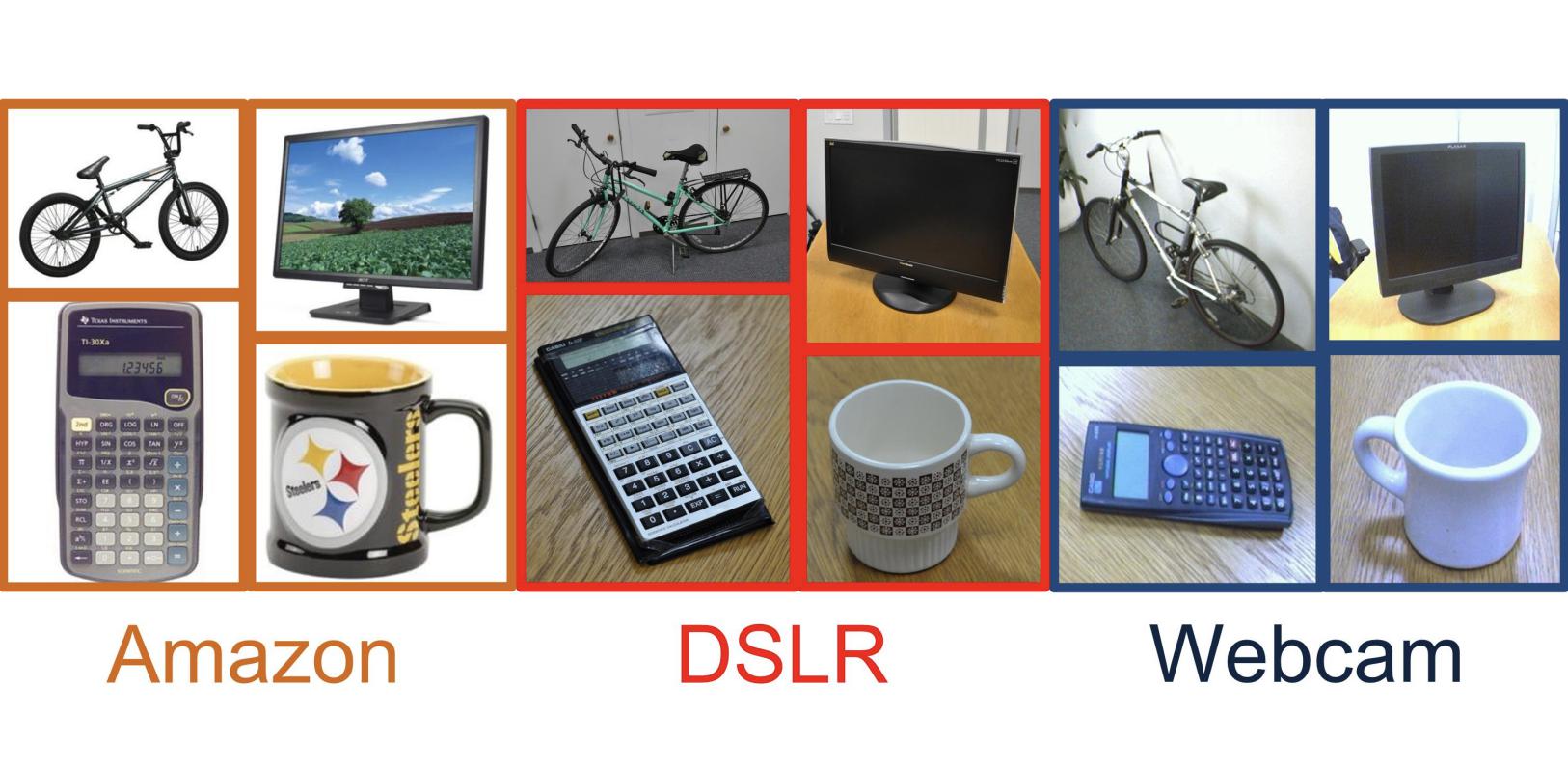}
    \caption{Different domains in dataset \textbf{Office-31}.}
    \label{fig:office}
\end{figure}

The dataset \textbf{Office-31} is used to test the approaches. This dataset consists of three domains: Amazon (A) - 2817 images, Dslr (D) - 498 images and Webcam (W) - 795 images (see Figure \ref{fig:office}). 

The Table \ref{tab:features} below highlights various key features of the dataset, including the number of classes, image resolution, task and evaluation metric.

\begin{table}[h]
\centering
\caption{General information about the Office-31 dataset}
\label{tab:features}
\begin{tabular}{|l|l|}
\hline
\textbf{Feature}  & \textbf{Description}                     \\ \hline
Dataset Name      & Office-31                                \\ \hline
Purpose           & Domain adaptation and object recognition \\ \hline
Number of Classes & 31                                       \\ \hline
Domains           & Amazon, Webcam, DSLR                     \\ \hline
Image Resolution  & Varies across domains and images         \\ \hline
Image Format      & JPG                                     \\ \hline
Task              & Multi-class classification               \\ \hline
Evaluation Metric & Classification accuracy                  \\ \hline
\end{tabular}
\end{table}

The existence of dissimilar image quantities ensures us in the importance of utilizing one of the approaches discussed in the previous section in order to avoid any information loss. In the first approach, where the smaller domain is concatenated, a batch size of 32 or 64 is utilized for all experiments. The second approach takes into account the size of each domain, and as a result, the batch sizes are utilized in the experiments according to the Table \ref{tab:batchsizes}:

\begin{table}[H]
\centering
\caption{The table of batch sizes is organized such that the numerical values in the first column correspond to the first letter, the second one to the second letter.}
\label{tab:batchsizes}
\begin{tabular}{|p{1.5cm}|p{1.5cm}|p{1.5cm}|}
\hline
   & \multicolumn{2}{l|}{Batch size} \\ \hline
\textit{AD} & 45   & 8   \\ \hline
\textit{AW} & 45   & 13  \\ \hline
\textit{DA} & 8    & 45  \\ \hline
\textit{DW} & 20   & 32  \\ \hline
\textit{WA} & 13   & 45  \\ \hline
\textit{WD} & 32   & 20  \\ \hline
\end{tabular}
\end{table}

For the all methods, two kinds of optimizers are used: SGD and Adam. However, the second one shows worse results with default parameters than SGD with lr = 0.001, momentum = 0.9, weight~decay = 0.0005. The CustomLRScheduler and CosineAnnealingLR are both used as schedulers, but it has been found that the model performs better when using the second one. Thus, all the following results have been obtained using the CosineAnnealingLR scheduler. Furthermore, all methods give the best results with an approach using a different number of images in each batch. As a result, this approach will be assumed by default, unless otherwise stated. For each two domains, at least three experiments were conducted for all methods, and the best results were selected.\\

Let's start with the first method -- \textbf{Source only}. Here, the model is trained on the source domain and then tested on the target domain. The obtained results are shown in Table \ref{tab:source} (at the end of the 60th epoch).

\begin{table}[h]
\centering
\caption{Accuracy on Office-31 for the \textbf{Source only} method.}
\label{tab:source}
\begin{tabular}{|r|r|r|r|r|r|r|}
\hline
\multicolumn{1}{|l|}{Source} & \multicolumn{1}{l|}{A$\rightarrow$D} & \multicolumn{1}{l|}{A$\rightarrow$W} & \multicolumn{1}{l|}{D$\rightarrow$W} & \multicolumn{1}{l|}{D$\rightarrow$A} & \multicolumn{1}{l|}{W$\rightarrow$D} & \multicolumn{1}{l|}{W$\rightarrow$A} \\ \hline
1 & 81.46 & 76.43 & 95.96 & 60.33 & 99.58 & 64.45 \\ \hline
2 & 81.04 & 76.04 & 95.7 & 60.09 & 99.37 & 64.17 \\ \hline
3 & 80.62 & 76.04 & 95.7 & 59.38 & 99.37 & 64.13 \\ \hline
\multicolumn{1}{|l|}{\textbf{average}} & \textbf{81.04} & \textbf{76.17} & \textbf{95.79} & \textbf{59.93} & \textbf{99.44} & \textbf{64.25} \\ \hline
\end{tabular}
\end{table}

\textbf{Dann} is an architecture that consists not only of a feature extractor and label predictor, but also a domain classifier. This domain classifier helps to identify the domain of the input data and allows the model to learn domain-invariant features. The Table \ref{tab:dann} below clearly demonstrates that the results for each of the two domains are superior to those obtained using the simple \textbf{Source only} method. The results are obtained at the end of the 60th epoch.

\begin{table}[h]
\centering
\caption{Accuracy on Office-31 for the \textbf{Dann} method.}
\label{tab:dann}
\begin{tabular}{|r|r|r|r|r|r|r|}
\hline
\multicolumn{1}{|l|}{Dann} & \multicolumn{1}{l|}{A$\rightarrow$D} & \multicolumn{1}{l|}{A$\rightarrow$W} & \multicolumn{1}{l|}{D$\rightarrow$W} & \multicolumn{1}{l|}{D$\rightarrow$A} & \multicolumn{1}{l|}{W$\rightarrow$D} & \multicolumn{1}{l|}{W$\rightarrow$A} \\ \hline
1 & 83.13 & 78.91 & 96.35 & 64.35 & 100 & 65.38 \\ \hline
2 & 82.92 & 78.65 & 95.96 & 63.88 & 100 & 64.91 \\ \hline
3 & 82.71 & 78.65 & 95.83 & 63.92 & 99.79 & 64.74 \\ \hline
\multicolumn{1}{|l|}{\textbf{average}} & \textbf{82.92} & \textbf{78.74} & \textbf{96.05} & \textbf{64.05} & \textbf{99.93} & \textbf{65.01} \\ \hline
\end{tabular}
\end{table}

\textbf{Mstn} method is a complication of \textbf{Dann} by adding a semantic loss. To get this loss, we add centroids for each class and utilize the algorithm described in the section 1.2.2. In the Table \ref{tab:mstn}, you can see the results that are acquired at the end of the 60th epoch. The quality of the results tends to suffer due to the significant influence of randomness.  The selection of pictures that are included in a batch determines the movements of the centroids, ultimately influencing the overall quality to a significant extent.

\begin{table}[h]
\centering
\caption{Accuracy on Office-31 for the \textbf{Mstn} method.}
\label{tab:mstn}
\begin{tabular}{|r|r|r|r|r|r|r|}
\hline
\multicolumn{1}{|l|}{Mstn} & \multicolumn{1}{l|}{A$\rightarrow$D} & \multicolumn{1}{l|}{A$\rightarrow$W} & \multicolumn{1}{l|}{D$\rightarrow$W} & \multicolumn{1}{l|}{D$\rightarrow$A} & \multicolumn{1}{l|}{W$\rightarrow$D} & \multicolumn{1}{l|}{W$\rightarrow$A} \\ \hline
1 & 77.71 & 72.53 & 92.06 & 33.38 & 99.79 & 51.07 \\ \hline
2 & 76.88 & 72.4 & 91.8 & 32.1 & 99.79 & 48.58 \\ \hline
3 & 76.67 & 71.48 & 91.41 & 34.09 & 99.79 & 48.65 \\ \hline
\multicolumn{1}{|l|}{\textbf{average}} & \textbf{77.09} & \textbf{72.14} & \textbf{91.76} & \textbf{33.19} & \textbf{99.79} & \textbf{49.43} \\ \hline
\end{tabular}
\end{table}

\begin{figure}[H]
    \centering
    \includegraphics[width=0.8\textwidth]{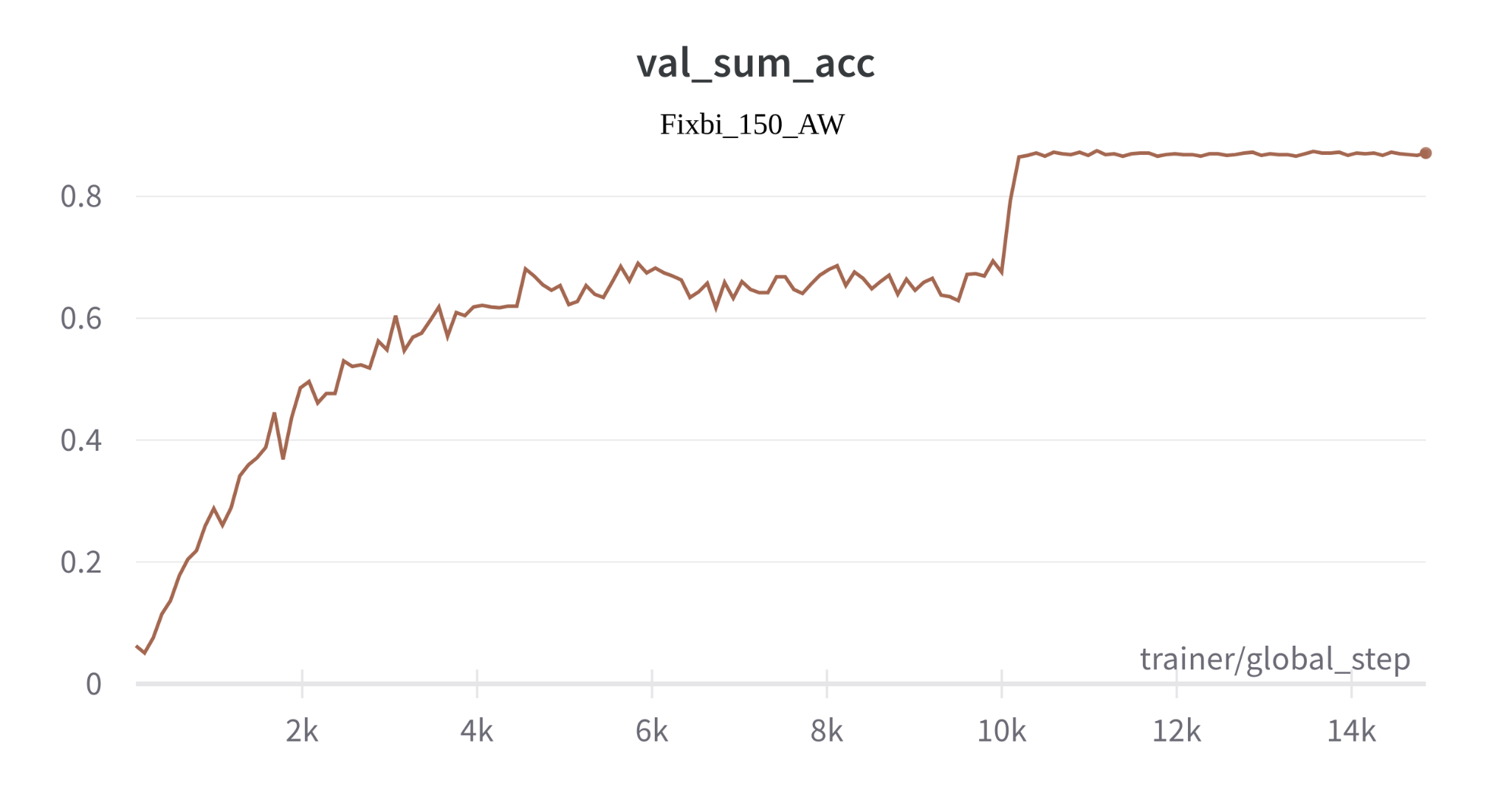}
    \caption{The total accuracy of \textbf{Fixbi} model on Amazon and Webcam domains.}
    \label{fig:fixbi_total}
\end{figure}

\textbf{Fixbi} is a method that addresses the domain adaptation problem by training two neural networks that can help each other. As it is described in the article \cite{na2021fixbi}, we define $\lambda_{sd} = 0.7$ and $\lambda_{td} = 0.3$. In this method, we cannot use the approach with different batch sizes due to the need to mix up images from source and target domain. Therefore, the second approach with concatenation is utilized. The model is trained for a combined duration of 150 epochs, with the first 100 epochs designated as the warm-up period. It is important to note that 150 epochs are used, not 200, because after the warm-up period the validation score stabilizes and almost does not change. After the warm-up period, $L_{bim}$ starts to be applied, which leads to a critical changing in the total accuracy. The sudden improvement in accuracy can occur in either a positive or negative direction, and is often heavily influenced by randomness. One possible explanation for this phenomenon is that the model may have already found a local minimum prior to the introduction of $L_{bim}$, and the application of $L_{bim}$ causes a sudden shift in gradients that propels the model out of the current minimum and into a new one. Depending on the new minimum, this can result in either an improvement or a deterioration in the model's performance. As we can see in the Figure \ref{fig:fixbi_total}, for Amazon (source) and Webcam (target) domains this method gives significant increase in accuracy, while for DSLR (source) and Amazon (target) it shows the worst results. 

\begin{figure}[h]
  \centering
  \begin{minipage}[b]{0.49\textwidth}
    \includegraphics[width=\textwidth]{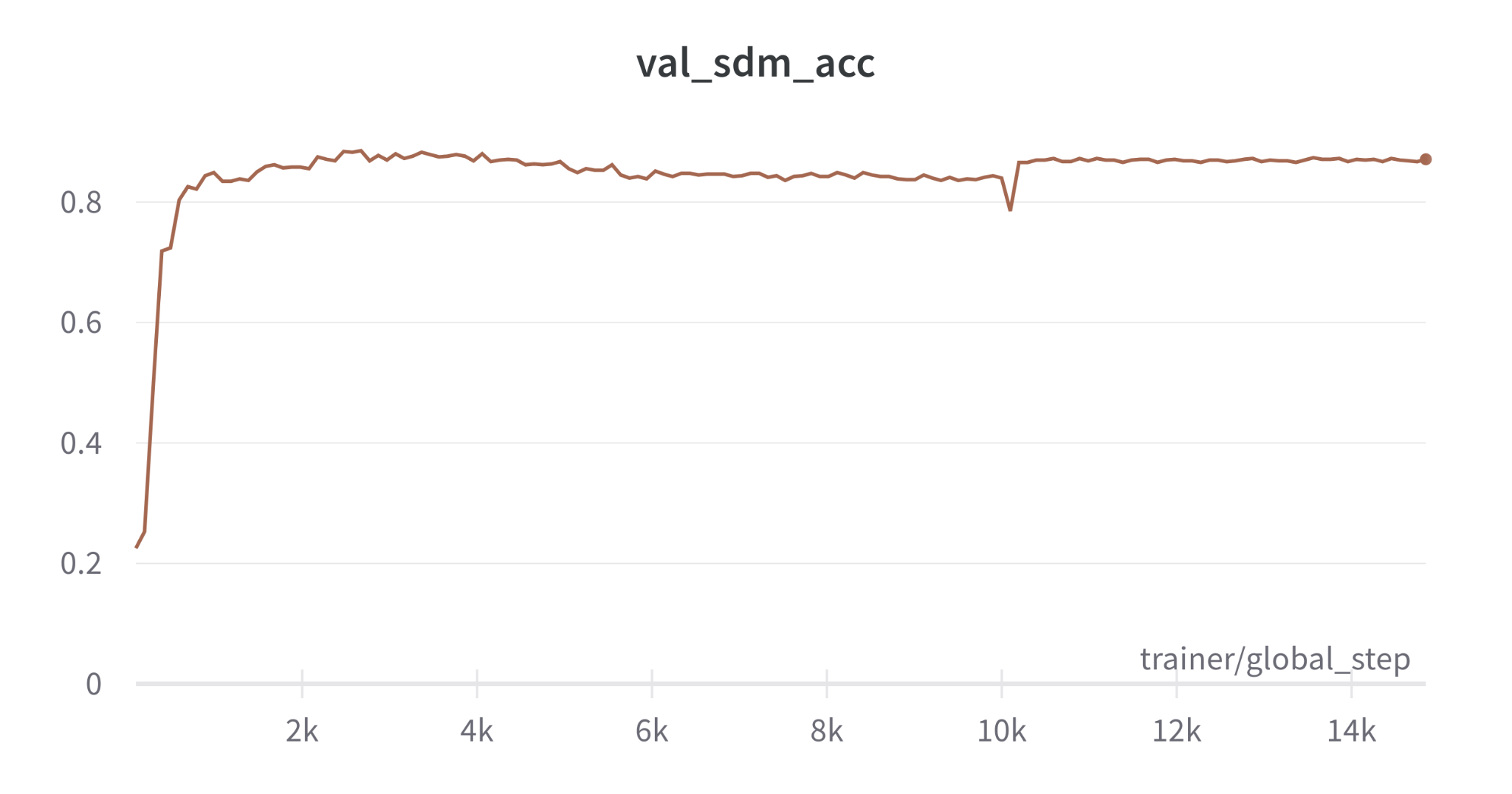}
  \end{minipage}
  \hfill
  \begin{minipage}[b]{0.49\textwidth}
    \includegraphics[width=\textwidth]{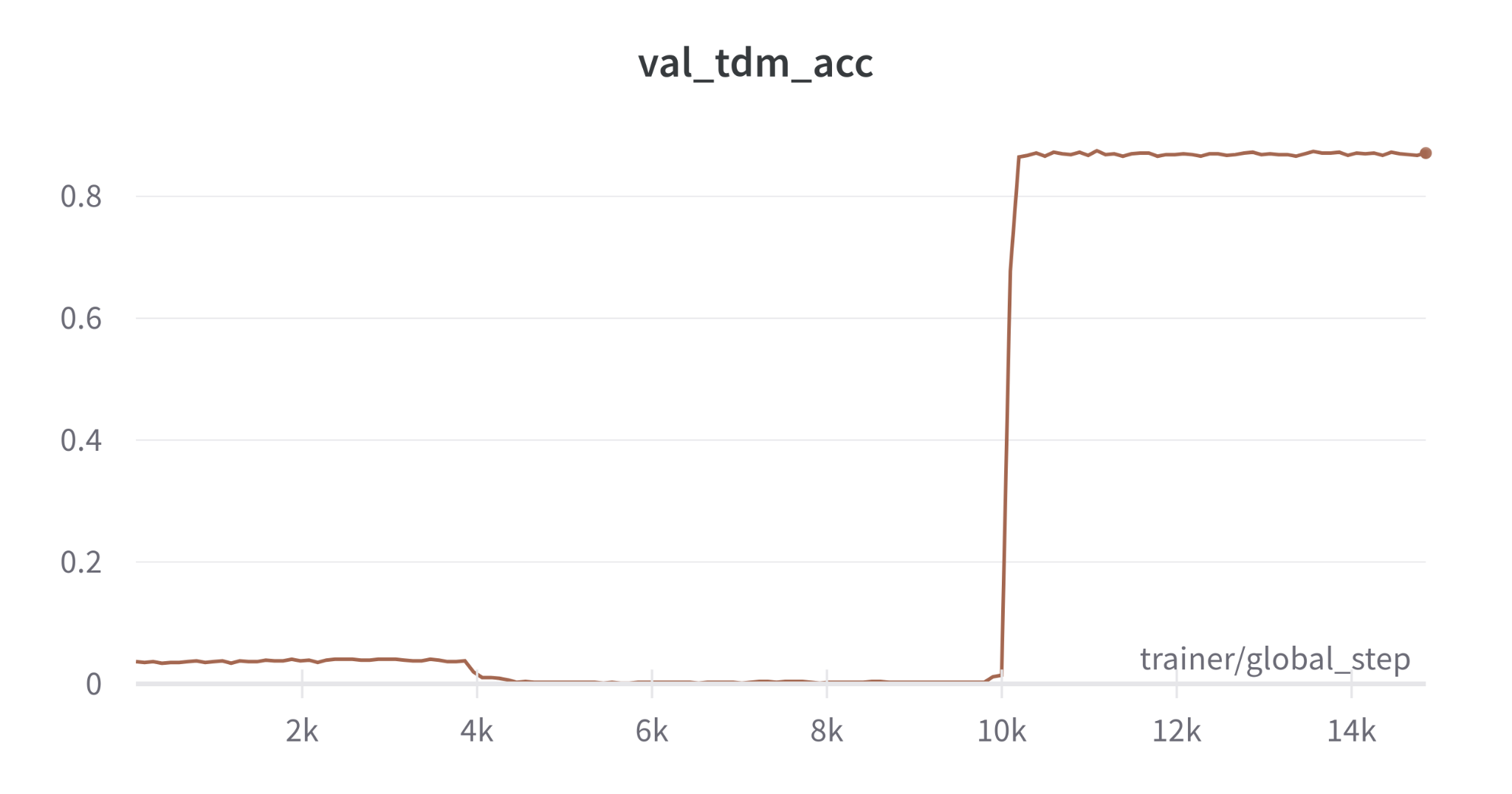}
  \end{minipage}
  \caption{The separate accuracy of \textbf{Fixbi} “source-dominant model” (SDM) on the left side and “target-dominant model” (TDM) on the right side.}
  \label{fig:sdm_tdm}
\end{figure}

In the Figure \ref{fig:sdm_tdm}, you can see the separate accuracy of “source-dominant model” (SDM) and “target-dominant model” (TDM) in case of Amazon (source) and Webcam (target) domains.

The results for each two domains of Office-31 dataset for \textbf{Fixbi} method are shown in Table \ref{tab:fixbi}.\\
\begin{table}[h]
\centering
\caption{Accuracy on Office-31 for the \textbf{Fixbi} method.}
\label{tab:fixbi}
\begin{tabular}{|r|r|r|r|r|r|r|}
\hline
\multicolumn{1}{|l|}{Fixbi} & \multicolumn{1}{l|}{A$\rightarrow$D} & \multicolumn{1}{l|}{A$\rightarrow$W} & \multicolumn{1}{l|}{D$\rightarrow$W} & \multicolumn{1}{l|}{D$\rightarrow$A} & \multicolumn{1}{l|}{W$\rightarrow$D} & \multicolumn{1}{l|}{W$\rightarrow$A} \\ \hline
1 & 76.88 & 87.11 & 94.66 & 23.19 & 94.38 & 30.15 \\ \hline
2 & 70.1 & 86.72 & 90.89 & 15.23 & 94.38 & 26.03 \\ \hline
3 & 77.08 & 81.77 & 88.54 & 16.01 & 92.92 & 20.63 \\ \hline
4 & 73.96 & 81.38 & 92.78 & 17.8 & 92.88 & 23.45 \\ \hline
\multicolumn{1}{|l|}{\textbf{average}} & \textbf{74.51} & \textbf{84.25} & \textbf{91.72} & \textbf{18.06} & \textbf{93.64} & \textbf{25.07} \\ \hline
\end{tabular}
\end{table}

The \textbf{Fixbi} method was selected for modification, wherein the ratios for SDM and TDM were adjusted, and a domain classifier was added for mixup images. This modified method is named \textbf{DannFixbi}. $\lambda_{sd}$ and $\lambda_{td}$ are set as $0.9$ and $0.7$, respectively. As it was mentioned before, it is used one of the two approaches described in "Contribution" section with domain classifiers for mixup images or separately for each domain. The Table \ref{tab:dann_fixbi} indicates that certain domains exhibit an increase in accuracy as a result of these changes.

\begin{table}[h]
\centering
\caption{Accuracy on Office-31 for the \textbf{DannFixbi} method.}
\label{tab:dann_fixbi}
\begin{tabular}{|r|r|r|r|r|r|r|}
\hline
\multicolumn{1}{|l|}{DannFixbi} & \multicolumn{1}{l|}{A$\rightarrow$D} & \multicolumn{1}{l|}{A$\rightarrow$W} & \multicolumn{1}{l|}{D$\rightarrow$W} & \multicolumn{1}{l|}{D$\rightarrow$A} & \multicolumn{1}{l|}{W$\rightarrow$D} & \multicolumn{1}{l|}{W$\rightarrow$A} \\ \hline
1 & 86.87 & 85.81 & 97.53 & 65.27 & 99.37 & 63.04 \\ \hline
2 & 86.46 & 86.07 & 97.35 & 64.81 & 98.98 & 63.29 \\ \hline
3 & 87.29 & 86.07 & 97.44 & 63.53 & 99.17 & 62.93 \\ \hline
\multicolumn{1}{|l|}{\textbf{average}} & \textbf{86.87} & \textbf{85.98} & \textbf{97.44} & \textbf{64.54} & \textbf{99.17} & \textbf{63.09} \\ \hline
\end{tabular}
\end{table}

Table \ref{tab:all} presents all the obtained results. The \textbf{DannFixbi} method yields the highest accuracy for the A$\rightarrow$D, A$\rightarrow$W, D$\rightarrow$A and D$\rightarrow$W tasks, while the \textbf{Dann} method achieves the best results for the W$\rightarrow$D, and W$\rightarrow$A tasks.

\begin{table}[H]
\centering
\caption{Accuracy on Office-31 for all methods.}
\label{tab:all}
\begin{tabular}{|r|r|r|r|r|r|r|}
\hline
\multicolumn{1}{|l|}{} & \multicolumn{1}{l|}{A$\rightarrow$D} & \multicolumn{1}{l|}{A$\rightarrow$W} & \multicolumn{1}{l|}{D$\rightarrow$W} & \multicolumn{1}{l|}{D$\rightarrow$A} & \multicolumn{1}{l|}{W$\rightarrow$D} & \multicolumn{1}{l|}{W$\rightarrow$A} \\ \hline
\textbf{Source} & 81.04 & 76.17 & 95.79 & 59.93 & 99.44 & 64.25 \\ \hline
\textbf{Dann} & 82.92 & 78.74 & 96.05 & 64.05 & \textbf{99.93} & \textbf{65.01} \\ \hline
\textbf{Mstn} & 77.09 & 72.14 & 91,76 & 33.19 & 99.79 & 49.43 \\ \hline
\textbf{Fixbi} & 74.51 & 84.25 & 91.72 & 18.06 & 93.64 & 25.07 \\ \hline
\textbf{DannFixbi} & \textbf{86.87} & \textbf{85.98} & \textbf{97.44} & \textbf{64.54} & 99.17 & 63.09 \\ \hline
\end{tabular}
\end{table}

Additionally, it is worth noting that an overall assessment of the  methods across all domains can be obtained by calculating the average accuracy (see Table \ref{tab:avg}).

\begin{table}[H]
\centering
\caption{Average accuracy across all domains}
\label{tab:avg}
\begin{tabular}{|r|r|}
\hline
                          & \multicolumn{1}{c|}{Avg} \\ \hline
\textbf{Source}           & 79.44                    \\ \hline
\textbf{Dann}             & 81.12                    \\ \hline
\textbf{Mstn} & 70.57                    \\ \hline
\textbf{Fixbi}            & 64.54                    \\ \hline
\textbf{DannFixbi}        & \textbf{82.85}           \\ \hline
\end{tabular}
\end{table}

To sum up, the new introduced method called \textbf{DannFixbi} outperforms all other methods in visual recognition. The \autoref{section: appendix A}  provides additional results for each domain, allowing for comparisons using the Wilcoxon signed-rank test to determine the statistical significance of the findings (see Tables \ref{tab:AD} -- \ref{tab:WA_wil}). The new method demonstrates statistically significant results in three tasks: A$\rightarrow$D, A$\rightarrow$W, and D$\rightarrow$W, while outperforming all other methods on average (Tables \ref{tab:all_avg} -- \ref{tab:tot_avg}).

\newpage
\section{Conclusion and Perspectives}

In this thesis, the focus was on exploring and implementing methods related to unsupervised domain adaptation. The Office-31 dataset was utilized for evaluating these methods and conducting a comprehensive comparison. The results obtained from the experiments were analyzed, leading to the development of the new \textbf{DannFixbi} method that demonstrated the best performance compared to all the other methods presented. 

\smallskip

The Office-31 dataset provided a suitable benchmark for evaluating the effectiveness of various unsupervised domain adaptation techniques. By conducting experiments on this dataset, the performance of different methods could be objectively assessed and compared. The analysis of the results shows the strengths and weaknesses of each method, allowing a deeper understanding of their capabilities. 

\smallskip

Based on the comparative analysis, it was observed that the newly developed method showcased the best results among all the presented methods. The success of the new method can be attributed to its ability to leverage the strengths of existing techniques. By combining the back propagation method with domain classifiers and applying the Fixbi approach, it is possible to identify common features in different domains and share knowledge and insights between networks. This collaborative approach to learning has led to higher performance and increased the overall effectiveness of the method.

\smallskip

Overall, this thesis contributes to the field of unsupervised domain adaptation by providing an analysis of existing methods, introducing a new approach, and demonstrating the potential for improving visual recognition tasks across different domains. The results of this study open up opportunities for further study and development of advanced methods in the field of domain adaptation.

\smallskip

By addressing the challenge of distribution mismatch between the labeled and unlabeled data, we can note that advances in domain adaptation can significantly benefit other related domains specially semi-supervised learning. One line of research would be to study the generalization performance of semi-supervised learning models that have been studied under the cluster assumption \cite{10.5555/3241691.3241708}. Indeed, by explicitly considering the differences between the source and target domains, domain adaptation techniques can enhance the model's ability to adapt to new, unseen data in the target domain and can hence provide strategies to handle domain shift and improve the generalization performance of the semi-supervised learning model. Furthermore,  by reducing the distribution mismatch between labeled and unlabeled data, domain adaptation methods can enable semi-supervised learning algorithms to leverage the unlabeled data more effectively. Moreover, domain adaptation methods often focus on learning robust representations that are less sensitive to noise and domain shifts. By leveraging such robust representations, semi-supervised learning algorithms can become more resilient to label noise and improve their accuracy even with limited labeled data. Domain adaptation is essentially a form of transfer learning, where knowledge learned from a source domain is transferred to a target domain. By studying domain adaptation, we can gain insights into transfer learning techniques that can be beneficial for semi-supervised learning scenarios like ranking \cite{10.5555/2034161.2034191}. These techniques can help leverage knowledge from a labeled source domain to improve the performance of a semi-supervised learning model in the target domain.
\newpage

\bibliographystyle{unsrt}
\bibliography{References.bib}

\begin{thebibliography}{10}

\bibitem{redko2020survey}
Ievgen Redko, Emilie Morvant, Amaury Habrard, Marc Sebban, and Youn{\`e}s
  Bennani.
\newblock A survey on domain adaptation theory: learning bounds and theoretical
  guarantees.
\newblock {\em arXiv preprint arXiv:2004.11829}, 2020.

\bibitem{wilson2020survey}
Garrett Wilson and Diane~J Cook.
\newblock A survey of unsupervised deep domain adaptation.
\newblock {\em ACM Transactions on Intelligent Systems and Technology (TIST)},
  11(5):1--46, 2020.

\bibitem{Amini00}
Massih-Reza Amini.
\newblock Interactive learning for text summarization.
\newblock In {\em Proceedings of the PKDD/MLTIA Workshop on Machine Learning
  and Textual Information Access}, Lyon - France, 2000.

\bibitem{Pessiot09}
Jean-François Pessiot, Young-Min Kim, Massih-Reza Amini, and Patrick
  Gallinari.
\newblock Improving document clustering in a learned concept space.
\newblock {\em Information Processing \& Management}, 46(2):180--192, 2010.

\bibitem{zhang2021survey}
Youshan Zhang.
\newblock A survey of unsupervised domain adaptation for visual recognition,
  2021.

\bibitem{wang2018theoretical}
Zirui Wang.
\newblock Theoretical guarantees of transfer learning.
\newblock {\em arXiv preprint arXiv:1810.05986}, 2018.

\bibitem{ganin2015unsupervised}
Yaroslav Ganin and Victor Lempitsky.
\newblock Unsupervised domain adaptation by backpropagation.
\newblock In {\em International conference on machine learning}, pages
  1180--1189. PMLR, 2015.

\bibitem{xie2018learning}
Shaoan Xie, Zibin Zheng, Liang Chen, and Chuan Chen.
\newblock Learning semantic representations for unsupervised domain adaptation.
\newblock In {\em International conference on machine learning}, pages
  5423--5432. PMLR, 2018.

\bibitem{na2021fixbi}
Jaemin Na, Heechul Jung, Hyung~Jin Chang, and Wonjun Hwang.
\newblock Fixbi: Bridging domain spaces for unsupervised domain adaptation.
\newblock In {\em Proceedings of the IEEE/CVF Conference on Computer Vision and
  Pattern Recognition}, pages 1094--1103, 2021.

\bibitem{gu2020spherical}
Xiang Gu, Jian Sun, and Zongben Xu.
\newblock Spherical space domain adaptation with robust pseudo-label loss.
\newblock In {\em Proceedings of the IEEE/CVF Conference on Computer Vision and
  Pattern Recognition}, pages 9101--9110, 2020.

\bibitem{stojanov2021domain}
Petar Stojanov, Zijian Li, Mingming Gong, Ruichu Cai, Jaime Carbonell, and Kun
  Zhang.
\newblock Domain adaptation with invariant representation learning: What
  transformations to learn?
\newblock {\em Advances in Neural Information Processing Systems},
  34:24791--24803, 2021.

\bibitem{zou2018unsupervised}
Yang Zou, Zhiding Yu, BVK Kumar, and Jinsong Wang.
\newblock Unsupervised domain adaptation for semantic segmentation via
  class-balanced self-training.
\newblock In {\em Proceedings of the European conference on computer vision
  (ECCV)}, pages 289--305, 2018.

\bibitem{richter2016playing}
Stephan~R Richter, Vibhav Vineet, Stefan Roth, and Vladlen Koltun.
\newblock Playing for data: Ground truth from computer games.
\newblock In {\em Computer Vision--ECCV 2016: 14th European Conference,
  Amsterdam, The Netherlands, October 11-14, 2016, Proceedings, Part II 14},
  pages 102--118. Springer, 2016.

\bibitem{cordts2016cityscapes}
Marius Cordts, Mohamed Omran, Sebastian Ramos, Timo Rehfeld, Markus Enzweiler,
  Rodrigo Benenson, Uwe Franke, Stefan Roth, and Bernt Schiele.
\newblock The cityscapes dataset for semantic urban scene understanding.
\newblock In {\em Proceedings of the IEEE conference on computer vision and
  pattern recognition}, pages 3213--3223, 2016.

\bibitem{saenko2010adapting}
Kate Saenko, Brian Kulis, Mario Fritz, and Trevor Darrell.
\newblock Adapting visual category models to new domains.
\newblock In {\em Computer Vision--ECCV 2010: 11th European Conference on
  Computer Vision, Heraklion, Crete, Greece, September 5-11, 2010, Proceedings,
  Part IV 11}, pages 213--226. Springer, 2010.

\bibitem{long2017deep}
Mingsheng Long, Han Zhu, Jianmin Wang, and Michael~I Jordan.
\newblock Deep transfer learning with joint adaptation networks.
\newblock In {\em International conference on machine learning}, pages
  2208--2217. PMLR, 2017.

\bibitem{venkateswara2017deep}
Hemanth Venkateswara, Jose Eusebio, Shayok Chakraborty, and Sethuraman
  Panchanathan.
\newblock Deep hashing network for unsupervised domain adaptation.
\newblock In {\em Proceedings of the IEEE conference on computer vision and
  pattern recognition}, pages 5018--5027, 2017.

\bibitem{peng2017visda}
Xingchao Peng, Ben Usman, Neela Kaushik, Judy Hoffman, Dequan Wang, and Kate
  Saenko.
\newblock Visda: The visual domain adaptation challenge.
\newblock {\em arXiv preprint arXiv:1710.06924}, 2017.

\bibitem{peng2019moment}
Xingchao Peng, Qinxun Bai, Xide Xia, Zijun Huang, Kate Saenko, and Bo~Wang.
\newblock Moment matching for multi-source domain adaptation.
\newblock In {\em Proceedings of the IEEE International Conference on Computer
  Vision}, pages 1406--1415, 2019.

\bibitem{he2016deep}
Kaiming He, Xiangyu Zhang, Shaoqing Ren, and Jian Sun.
\newblock Deep residual learning for image recognition.
\newblock In {\em Proceedings of the IEEE conference on computer vision and
  pattern recognition}, pages 770--778, 2016.

\bibitem{10.5555/3241691.3241708}
Yury Maximov, Massih-Reza Amini, and Zaid Harchaoui.
\newblock Rademacher complexity bounds for a penalized multi-class
  semi-supervised algorithm.
\newblock {\em Journal of Artificial Intelligence Research}, 61(1):761–786,
  jan 2018.

\bibitem{10.5555/2034161.2034191}
Nicolas Usunier, Massih-Reza Amini, and Cyril Goutte.
\newblock Multiview semi-supervised learning for ranking multilingual
  documents.
\newblock In {\em Proceedings of the 2011 European Conference on Machine
  Learning and Knowledge Discovery in Databases - Volume Part III}, ECML
  PKDD'11, page 443–458, Berlin, Heidelberg, 2011. Springer-Verlag.

\end{thebibliography}
\newpage

\appendix 

\section{Appendix} \label{section: appendix A}

\textit{This appendix presents the results for each domain. In order to make comparisons, 15 experiments were conducted for each method within each area. The Wilcoxon rank test was employed to analyze and assess the performance of the methods. The best-performing method for each experiment is denoted by bold values and statistical significance is denoted with a star (*).}

\begin{table}[h]
\centering
\caption{Accuracy on Office-31 for the source Amazon and the target Dslr domains.}
\label{tab:AD}
\begin{tabular}{|p{2cm}|p{2cm}|p{2cm}|p{2cm}|p{2cm}|p{2cm}|}
\hline
   & A$\rightarrow$D & A$\rightarrow$D & A$\rightarrow$D & A$\rightarrow$D & A$\rightarrow$D \\ \hline
 & Source & Dann & Mstn & Fixbi & DannFixbi \\ \hline
1 & 81.67 & 83.13 & 77.71 & 76.88 & 86.87 \\ \hline
2 & 81.04 & 82.92 & 76.88 & 70.1 & 86.46 \\ \hline
3 & 81.04 & 82.71 & 76.67 & 77.08 & 87.29 \\ \hline
4 & 80.21 & 83.33 & 75.63 & 73.96 & 84.79 \\ \hline
5 & 79.79 & 82.71 & 75.21 & 83.54 & 84.38 \\ \hline
6 & 79.17 & 82.29 & 74.79 & 79.79 & 85 \\ \hline
7 & 79.97 & 82.08 & 73.33 & 76.67 & 85.21 \\ \hline
8 & 80.75 & 81.67 & 73.33 & 70.83 & 85.73 \\ \hline
9 & 81.07 & 81.25 & 78.98 & 82.08 & 86.61 \\ \hline
10 & 79.9 & 80.21 & 75.97 & 83.13 & 84.41 \\ \hline
11 & 79.12 & 82.36 & 76.14 & 81.04 & 85.58 \\ \hline
12 & 79.47 & 82.29 & 76.83 & 82.5 & 85.42 \\ \hline
13 & 80.17 & 81.46 & 74.79 & 79.58 & 86.4 \\ \hline
14 & 81.51 & 82.5 & 76.68 & 73.75 & 85.08 \\ \hline
15 & 80.11 & 81.67 & 74.48 & 71.25 & 84.22 \\ \hline 
avg & 80.33 & 82.17 & 75.83 & 77.48 & \textbf{85.56}$^*$ \\ \hline
\end{tabular}
\end{table}

\begin{table}[h]
\centering
\caption{Wilcoxon signed-rank test for the source Amazon and the target Dslr domains.}
\label{tab:AD_wil}
\begin{tabular}{|l|l|}
\hline
 & A$\rightarrow$D \\ \hline
Source and DannFixbi & \textbf{$p(x > 3.4078) = \textbf{0.0117}$ $\leq$ 0.025} \\ \hline
Dann and DannFixbi & \textbf{$p(x > 3.4078) = \textbf{0.0117} $ $\leq$ 0.025} \\ \hline
Mstn and DannFixbi & \textbf{$p(x > 3.4078) = \textbf{0.0117}$ $\leq$ 0.025} \\ \hline
Fixbi and DannFixbi & \textbf{$p(x > 3.4078) = \textbf{0.0117}$ $\leq$ 0.025} \\ \hline
\end{tabular}
\end{table}

\begin{table}[H]
\centering
\caption{Accuracy on Office-31 for the source Amazon and the target Webcam domains.}
\label{tab:AW}
\begin{tabular}{|p{2cm}|p{2cm}|p{2cm}|p{2cm}|p{2cm}|p{2cm}|}
\hline
   & A$\rightarrow$W & A$\rightarrow$W & A$\rightarrow$W & A$\rightarrow$W & A$\rightarrow$W \\ \hline
 & Source & Dann & Mstn & Fixbi & DannFixbi \\ \hline
1 & 76.43 & 78.91 & 72.53 & 87.11 & 85.81 \\ \hline
2 & 76.04 & 78.65 & 72.4 & 86.72 & 86.07 \\ \hline
3 & 76.04 & 78.65 & 71.48 & 81.77 & 86.07 \\ \hline
4 & 76.45 & 80.23 & 70.23 & 81.38 & 83.07 \\ \hline
5 & 76.28 & 79.31 & 71.11 & 83.58 & 84.64 \\ \hline
6 & 75.44 & 79.15 & 73.81 & 80.57 & 85.29 \\ \hline
7 & 76.88 & 80.05 & 71.55 & 85.1 & 82.81 \\ \hline
8 & 75.92 & 80.66 & 69.55 & 81.68 & 85.55 \\ \hline
9 & 77.22 & 78.79 & 70.54 & 80.28 & 84.51 \\ \hline
10 & 76.09 & 79.05 & 71.21 & 82.78 & 85.42 \\ \hline
11 & 78.7 & 79.83 & 69.36 & 84.21 & 84.9 \\ \hline
12 & 75.67 & 78.81 & 70.28 & 81.08 & 82.29 \\ \hline
13 & 75.9 & 80.47 & 71.67 & 79.27 & 83.33 \\ \hline
14 & 76.71 & 78.2 & 71.72 & 83.44 & 82.68 \\ \hline
15 & 77.65 & 78.77 & 72.1 & 81.11 & 81.64 \\ \hline
avg & 76.49 & 79.30 & 71.30 & 82.67 & \textbf{84.27}$^*$ \\ \hline
\end{tabular}
\end{table}

\begin{table}[H]
\centering
\caption{Wilcoxon signed-rank test for the source Amazon and the target Webcam domains.}
\label{tab:AW_wil}
\begin{tabular}{|l|l|}
\hline
 & A$\rightarrow$D \\ \hline
Source and DannFixbi & \textbf{$p(x > 3.4078) = \textbf{0.0117}$ $\leq$ 0.025} \\ \hline
Dann and DannFixbi & \textbf{$p(x > 3.4078) = \textbf{0.0117} $ $\leq$ 0.025} \\ \hline
Mstn and DannFixbi & \textbf{$p(x > 3.4078) = \textbf{0.0117}$ $\leq$ 0.025} \\ \hline
Fixbi and DannFixbi & \textbf{$p(x > 2.1583) = \textbf{0.0117}$ $\leq$ 0.025} \\ \hline
\end{tabular}
\end{table} 

\begin{table}[H]
\centering
\caption{Accuracy on Office-31 for the source Dslr and the target Webcam domains.}
\label{tab:DW}
\begin{tabular}{|p{2cm}|p{2cm}|p{2cm}|p{2cm}|p{2cm}|p{2cm}|}
\hline
   & D$\rightarrow$W & D$\rightarrow$W & D$\rightarrow$W & D$\rightarrow$W & D$\rightarrow$W \\ \hline
 & Source & Dann & Mstn & Fixbi & DannFixbi \\ \hline
1 & 95.96 & 96.35 & 92.06 & 94.66 & 97.53 \\ \hline
2 & 95.7 & 95.96 & 91.8 & 90.89 & 97.35 \\ \hline
3 & 95.7 & 95.83 & 91.41 & 88.54 & 97.44 \\ \hline
4 & 97.01 & 94.92 & 93.49 & 92.78 & 96.74 \\ \hline
5 & 94.79 & 96.48 & 93.1 & 93.1 & 96.88 \\ \hline
6 & 96.88 & 94.4 & 91.67 & 94.4 & 97.27 \\ \hline
7 & 95.44 & 95.96 & 92.32 & 91.69 & 97.4 \\ \hline
8 & 94.92 & 96.22 & 92.19 & 93.53 & 96.61 \\ \hline
9 & 96.74 & 96.35 & 95.44 & 92.29 & 97.14 \\ \hline
10 & 95.96 & 95.05 & 94.27 & 94.42 & 97.01 \\ \hline
11 & 95.05 & 94.79 & 91.02 & 96.88 & 95.7 \\ \hline
12 & 95.7 & 96.61 & 93.23 & 91.92 & 97.53 \\ \hline
13 & 97.53 & 95.57 & 92.19 & 92.55 & 96.09 \\ \hline
14 & 95.18 & 97.01 & 90.23 & 93.53 & 96.35 \\ \hline
15 & 96.61 & 96.09 & 95.18 & 94.82 & 95.7 \\ \hline
avg & 95.94 & 95.84 & 92.64 & 93.07 & \textbf{96.85}$^*$ \\ \hline
\end{tabular}
\end{table}

\begin{table}[H]
\centering
\caption{Wilcoxon signed-rank test for the source Dslr and the target Webcam domains.}
\label{tab:DW_wil}
\begin{tabular}{|l|l|}
\hline
 & D$\rightarrow$W \\ \hline
Source and DannFixbi & \textbf{$p(x > 2.4422) = \textbf{0.0117}$ $\leq$ 0.025} \\ \hline
Dann and DannFixbi & \textbf{$p(x > 2.7262) = \textbf{0.0117} $ $\leq$ 0.025} \\ \hline
Mstn and DannFixbi & \textbf{$p(x > 3.3510) = \textbf{0.0117}$ $\leq$ 0.025} \\ \hline
Fixbi and DannFixbi & \textbf{$p(x > 3.2942) = \textbf{0.0117}$ $\leq$ 0.025} \\ \hline
\end{tabular}
\end{table} 

\begin{table}[H]
\centering
\caption{Accuracy on Office-31 for the source Dslr and the target Amazon domains.}
\label{tab:DA}
\begin{tabular}{|p{2cm}|p{2cm}|p{2cm}|p{2cm}|p{2cm}|p{2cm}|}
\hline
   & D$\rightarrow$A & D$\rightarrow$A & D$\rightarrow$A & D$\rightarrow$A & D$\rightarrow$A \\ \hline
 & Source & Dann & Mstn & Fixbi & DannFixbi \\ \hline
1 & 60.33 & 64.35 & 33.38 & 23.19 & 65.27 \\ \hline
2 & 60.09 & 63.88 & 32.1 & 15.23 & 64.81 \\ \hline
3 & 59.38 & 63.92 & 34.09 & 16.01 & 63.53 \\ \hline
4 & 58.03 & 62.6 & 36.01 & 17.8 & 62.43 \\ \hline
5 & 57.17 & 63.36 & 37.71 & 43.54 & 61.54 \\ \hline
6 & 58.98 & 62.31 & 38.6 & 25 & 61.93 \\ \hline
7 & 59.2 & 62.48 & 34.94 & 20.06 & 60.72 \\ \hline
8 & 60.23 & 61.11 & 33.1 & 25.5 & 61.26 \\ \hline
9 & 57.74 & 63.18 & 35.12 & 26.1 & 62.96 \\ \hline
10 & 58.1 & 63.94 & 34.2 & 29.58 & 64.88 \\ \hline
11 & 59.3 & 62.48 & 36.72 & 23.22 & 60.8 \\ \hline
12 & 60.72 & 62.9 & 32.81 & 20.7 & 62.18 \\ \hline
13 & 58.35 & 63.72 & 43.08 & 24.57 & 63.03 \\ \hline
14 & 59.17 & 63.6 & 39.38 & 23.08 & 64.7 \\ \hline
15 & 56.85 & 62.22 & 40.8 & 42.12 & 61.36 \\ \hline
avg & 58.91 & \textbf{63.07} & 36.14 & 25.05 & 62.76 \\ \hline
\end{tabular}
\end{table}

\begin{table}[H]
\centering
\caption{Wilcoxon signed-rank test for the source Dslr and the target Amazon domains.}
\label{tab:DA_wil}
\begin{tabular}{|l|l|}
\hline
 & D$\rightarrow$A \\ \hline
Source and Dann & \textbf{$p(x > 3.4078) = \textbf{0.0117}$ $\leq$ 0.025} \\ \hline
Mstn and Dann & \textbf{$p(x > 3.4078) = \textbf{0.0117}$ $\leq$ 0.025} \\ \hline
Fixbi and Dann & \textbf{$p(x > 3.4078) = \textbf{0.0117}$ $\leq$ 0.025} \\\hline
DannFixbi and Dann & \textbf{$p(x > 1.0859)= \textbf{0.1425} $ $>$ 0.1} \\ \hline
\end{tabular}
\end{table}

\begin{table}[H]
\centering
\caption{Accuracy on Office-31 for the source Webcam and the target Dslr domains.}
\label{tab:WD}
\begin{tabular}{|p{2cm}|p{2cm}|p{2cm}|p{2cm}|p{2cm}|p{2cm}|}
\hline
   & W$\rightarrow$D & W$\rightarrow$D & W$\rightarrow$D & W$\rightarrow$D & W$\rightarrow$D \\ \hline
 & Source & Dann & Mstn & Fixbi & DannFixbi \\ \hline
1 & 99.58 & 100 & 99.79 & 94.38 & 99.37 \\ \hline
2 & 99.37 & 100 & 99.79 & 94.38 & 98.98 \\ \hline
3 & 99.37 & 99.79 & 99.79 & 92.92 & 99.17 \\ \hline
4 & 99.79 & 100 & 99.37 & 92.88 & 99.37 \\ \hline
5 & 99.58 & 99.79 & 99.58 & 94.79 & 98.96 \\ \hline
6 & 99.37 & 100 & 99.17 & 93.12 & 99.17 \\ \hline
7 & 99.37 & 99.58 & 99.17 & 95 & 99.37 \\ \hline
8 & 99.58 & 99.58 & 99.58 & 92.5 & 99.58 \\ \hline
9 & 99.79 & 99.79 & 99.17 & 92.92 & 99.17 \\ \hline
10 & 99.58 & 100 & 99.79 & 92.08 & 99.58 \\ \hline
11 & 99.37 & 99.79 & 99.17 & 95.83 & 98.96 \\ \hline
12 & 99.79 & 99.58 & 99.37 & 94.79 & 99.17 \\ \hline
13 & 99.58 & 100 & 99.58 & 93.96 & 99.37 \\ \hline
14 & 99.58 & 99.79 & 99.17 & 91.87 & 99.58 \\ \hline
15 & 99.37 & 99.58 & 98.96 & 92.71 & 99.17 \\ \hline
avg & 99.54 & \textbf{99.82}$^*$ & 99.43 & 93.61 & 99.26 \\ \hline
\end{tabular}
\end{table}

\begin{table}[H]
\centering
\caption{Wilcoxon signed-rank test for the source Webcam and the target Dslr domains.}
\label{tab:WD_wil}
\begin{tabular}{|l|l|}
\hline
 & W$\rightarrow$D \\ \hline
Source and Dann & \textbf{$p(x > 2.8398) = \textbf{0.0117}$ $\leq$ 0.025} \\ \hline
Mstn and Dann & \textbf{$p(x > 3.2374) = \textbf{0.0117}$ $\leq$ 0.025} \\ \hline
Fixbi and Dann & \textbf{$p(x > 3.4078) = \textbf{0.0117}$ $\leq$ 0.025} \\\hline
DannFixbi and Dann & \textbf{$p(x > 3.3510)= \textbf{0.0117}$ $\leq$ 0.025} \\ \hline
\end{tabular}
\end{table} 

\begin{table}[H]
\centering
\caption{Accuracy on Office-31 for the source Webcam and the target Amazon domains.}
\label{tab:WA}
\begin{tabular}{|p{2cm}|p{2cm}|p{2cm}|p{2cm}|p{2cm}|p{2cm}|}
\hline
   & W$\rightarrow$A & W$\rightarrow$A & W$\rightarrow$A & W$\rightarrow$A & W$\rightarrow$A \\ \hline
 & Source & Dann & Mstn & Fixbi & DannFixbi \\ \hline
1 & 64.45 & 65.38 & 51.07 & 30.15 & 63.04 \\ \hline
2 & 64.17 & 64.91 & 48.58 & 26.03 & 63.29 \\ \hline
3 & 64.13 & 64.74 & 48.65 & 20.63 & 62.93 \\ \hline
4 & 62.96 & 64.7 & 50.6 & 23.45 & 62.32 \\ \hline
5 & 62.25 & 63.42 & 53.09 & 25.82 & 62.04 \\ \hline
6 & 63.14 & 64.91 & 49.68 & 23.69 & 61.97 \\ \hline
7 & 64.1 & 64.03 & 52.06 & 35.58 & 62.29 \\ \hline
8 & 61.75 & 63.81 & 52.49 & 27.7 & 61.54 \\ \hline
9 & 62.32 & 63.74 & 50.71 & 30.11 & 60.87 \\ \hline
10 & 62.46 & 64.1 & 51.81 & 24.08 & 61.93 \\ \hline
11 & 63 & 64.38 & 50.36 & 35.83 & 62.32 \\ \hline
12 & 62.82 & 62.57 & 49.43 & 30.29 & 62.22 \\ \hline
13 & 62.78 & 63.99 & 52.88 & 21.16 & 61.4 \\ \hline
14 & 63.39 & 62.82 & 55.15 & 25.92 & 60.9 \\ \hline
15 & 61.22 & 63.07 & 50.71 & 25.85 & 61.51 \\ \hline
avg & 63.00 & \textbf{64.04}$^*$ & 51.15 & 27.09 & 62.04 \\ \hline
\end{tabular}
\end{table}

\begin{table}[H]
\centering
\caption{Wilcoxon signed-rank test for the source Webcam and the target Amazon domains.}
\label{tab:WA_wil}
\begin{tabular}{|l|l|}
\hline
 & W$\rightarrow$A \\ \hline
Source and Dann & \textbf{$p(x > 3.0670) = \textbf{0.0117}$ $\leq$ 0.025} \\ \hline
Mstn and Dann & \textbf{$p(x > 3.4078) = \textbf{0.0117}$ $\leq$ 0.025} \\ \hline
Fixbi and Dann & \textbf{$p(x > 3.4078) = \textbf{0.0117}$ $\leq$ 0.025} \\\hline
DannFixbi and Dann & \textbf{$p(x > 3.4078)= \textbf{0.0117}$ $\leq$ 0.025} \\ \hline
\end{tabular}
\end{table} 

\begin{table}[H]
\caption{Accuracy on Office-31 for all methods.}
\label{tab:all_avg}
\hskip-1.2cm \begin{tabular}{|r|c|c|c|c|c|c|}
\hline
\multicolumn{1}{|c|}{} & \multicolumn{1}{c|}{A$\rightarrow$D} & \multicolumn{1}{c|}{A$\rightarrow$W} & \multicolumn{1}{c|}{D$\rightarrow$W} & \multicolumn{1}{c|}{D$\rightarrow$A} & \multicolumn{1}{c|}{W$\rightarrow$D} & \multicolumn{1}{c|}{W$\rightarrow$A} \\ \hline
\textbf{Source} & 80.33 $\pm$ 0.8 & 76.49 $\pm$ 0.8 & 95.94 $\pm$ 0.8 & 58.91 $\pm$ 1.2 & 99.54 $\pm$ 0.2 & 63.00 $\pm$ 0.9\\ \hline
\textbf{Dann} & 82.17 $\pm$ 0.8 & 79.3 $\pm$ 0.8 & 95.84 $\pm$ 0.7 & \textbf{63.07} $\pm$ 0.9& \textbf{99.82} $\pm$ 0.2 & \textbf{64.04} $\pm$ 0.8\\ \hline
\textbf{Mstn} & 75.83 $\pm$ 1.6 & 71.3 $\pm$ 1.2 & 92.64 $\pm$ 1.5 & 36.14 $\pm$ 3.2 & 99.43 $\pm$ 0.3 & 51.15 $\pm$ 1.8 \\ \hline
\textbf{Fixbi} & 77.48 $\pm$ 4.6& 82.67 $\pm$ 2.3 & 93.07 $\pm$ 2.0 & 25.05 $\pm$ 8.2 & 93.61 $\pm$ 1.2 & 27.09 $\pm$ 4.6 \\ \hline
\textbf{DannFixbi} & \textbf{85.56} $\pm$ 1.0 & \textbf{84.27} $\pm$ 1.5 & \textbf{96.85} $\pm$ 0.6 & 62.76 $\pm$ 1.6 & 99.26 $\pm$ 0.2 & 62.04 $\pm$ 0.7 \\ \hline
\end{tabular}
\end{table}

\begin{table}[H]
\centering
\caption{Average accuracy across all domains}
\label{tab:tot_avg}
\begin{tabular}{|r|r|}
\hline
                          & \multicolumn{1}{c|}{Avg} \\ \hline
\textbf{Source}           & 79.04                    \\ \hline
\textbf{Dann}             & 80.71                    \\ \hline
\textbf{Mstn} & 71.08                    \\ \hline
\textbf{Fixbi}            & 66.49                    \\ \hline
\textbf{DannFixbi}        & \textbf{81.79}           \\ \hline
\end{tabular}
\end{table}

\newpage

\end{document}